\documentclass{article}

\usepackage{PRIMEarxiv}

\usepackage[utf8]{inputenc} 
\usepackage[T1]{fontenc}    
\usepackage{hyperref}       
\usepackage{url}            
\usepackage{booktabs}       
\usepackage{amsfonts}       
\usepackage{nicefrac}       
\usepackage{microtype}      
\usepackage{lipsum}
\usepackage{fancyhdr}       
\usepackage{graphicx}       
\usepackage{xcolor}
\usepackage{amsmath}
\usepackage[at]{easylist}
\usepackage{subcaption}
\usepackage{multirow,adjustbox}
\graphicspath{{media/}}     

\title{Open Challenges and Opportunities in Federated Foundation Models Towards Biomedical Healthcare 
\thanks{\textit{\underline{Citation}}: 
\textbf{Authors. Title. Pages.... DOI:000000/11111.}} 
}

\author{
  Xingyu Li\\
  Department of Computer Science \\
  Tulane University \\
  New Orleans, LA\\
  \texttt{xli82@tulane.edu} \\
  \And
  Lu Peng \\
  Department of Computer Science \\
  Tulane University \\
  New Orleans, LA\\
  \texttt{lpeng3@tulane.edu} \\
  \And
  Yuping Wang \\
  Department of Biomedical Engineering\\
  Tulane University \\
  New Orleans, LA\\
  \texttt{wyp@tulane.edu} \\
  \AND
  Weihua Zhang \\
  School of Computer Science \\
  Fudan University\\
  Shanghai, China\\
  \texttt{zhangweihua@fudan.edu.cn} \\
}

\begin{document}
\maketitle

\begin{abstract}
  This survey explores the transformative impact of foundation models (FMs) in artificial intelligence, focusing on their integration with federated learning (FL) for advancing biomedical research. Foundation models such as ChatGPT, LLaMa, and CLIP, which are trained on vast datasets through methods including unsupervised pretraining, self-supervised learning, instructed fine-tuning, and reinforcement learning from human feedback, represent significant advancements in machine learning. These models, with their ability to generate coherent text and realistic images, are crucial for biomedical applications that require processing diverse data forms such as clinical reports, diagnostic images, and multimodal patient interactions.

  The incorporation of FL with these sophisticated models presents a promising strategy to harness their analytical power while safeguarding the privacy of sensitive medical data. This approach not only enhances the capabilities of FMs in medical diagnostics and personalized treatment but also addresses critical concerns about data privacy and security in healthcare. This survey reviews the current applications of FMs in federated settings, underscores the challenges, and identifies future research directions including scaling FMs, managing data diversity, and enhancing communication efficiency within FL frameworks. The objective is to encourage further research into the combined potential of FMs and FL, laying the groundwork for groundbreaking healthcare innovations.

\end{abstract}


\section{Introduction}
Foundation models (FMs) \cite{jeblick2023chatgpt, touvron2023llama} have risen to prominence as pivotal elements in the field of artificial intelligence \cite{lecun2015deep}. These models are distinguished by their deep learning architectures and a vast number of parameters, allowing them to excel in tasks ranging from text generation to video analysis—capabilities that surpass those of previous AI systems. FMs are developed using advanced training techniques, including unsupervised pretraining \cite{erhan2010does,caron2019unsupervised}, self-supervised training \cite{chen2021lottery,wang2021dense}, instructed fine-tuning \cite{liu2024visual}, and reinforcement human preference feedback \cite{christiano2017deep}. These methodologies equip them to generate coherent text and realistic images with unprecedented accuracy, showcasing their transformative potential across various domains.

The potential of foundation models extends far beyond mere technical capabilities. These models mark a significant paradigm shift in how we utilize artificial intelligence for cutting-edge scientific problem-solving. As versatile tools, they can be rapidly adapted and fine-tuned for specific tasks, eliminating the need to develop new models from the ground up. This adaptability is crucial in fields where processing limited datasets to extract meaningful insights is essential. It is particularly transformative in biomedical healthcare, where the efficacy of AI must be balanced with stringent data privacy considerations \cite{park2018machine, habehh2021machine,kaur2022trustworthy}. In this domain, foundation models not only enhance our analytical capabilities but also ensure that sensitive health information is handled with the utmost integrity, thereby aligning technological advancement with ethical standards.

Federated Learning (FL) \cite{mcmahan2017communication,li2023fedlga}, a method for training machine learning models across multiple decentralized devices or servers without exchanging local data samples, aligns well with the capabilities of foundation models in the biomedical healthcare sector. In this context, where data privacy and collaborative efforts are essential, FL enables the utilization of vast and varied datasets characteristic of the medical field while protecting sensitive patient information. By applying FL, foundation models can access and analyze extensive medical data without breaching privacy \cite{li2021survey,wei2020federated}, thus overcoming major obstacles in deploying AI technologies where data confidentiality is crucial. Existing applications of FL in conjunction with FMs typically involve training strategies that range from starting from scratch to prompt fine-tuning. FL enhances the application of FMs across both large language models and vision-language models, allowing for comprehensive and privacy-conscious analyses.

Integrating the privacy-preserving and decentralization features of FL with the robust, generalizable capabilities of FMs enables researchers to perform in-depth analyses using insights pooled from local datasets. This approach not only broadens the scope and accuracy of medical research but also complies with stringent data protection laws such as the General Data Protection Regulation (GDPR) \cite{li2019impact} in Europe and the Health Insurance Portability and Accountability Act (HIPAA) \cite{gostin2009beyond} in the United States. The potential for healthcare is profound, facilitating more personalized medicine where treatment plans are precisely tailored to individual genetic profiles, lifestyles, and medical histories. Additionally, FMs that are pre-trained or fine-tuned via federated learning on diverse datasets can reveal new biomarkers and therapeutic targets, thereby significantly pushing the boundaries of medical research and improving patient care. The synergy between federated learning and foundation models heralds a significant leap forward in the use of medical data, driving innovation in medical technologies while rigorously protecting patient privacy.

This paper presents a comprehensive survey of the latest advancements in foundation models and federated learning within the biomedical and healthcare sectors, highlighting their implementations and addressing the persistent challenges encountered in these fields. A notable application of these technologies involves the use of federated foundation models to train pre-trained vision-language models, such as FedClip \cite{lu2023fedclip}, which enhance both generalization and personalization in image classification tasks. Additionally, MedCLIP \cite{medclip2022} employs vision-text contrastive learning with 20K medical datasets to surpass current benchmarks in medical diagnostics. FedMed \cite{wu2020fedmed}, a tailored federated learning framework, effectively counters performance degradation in federated settings, facilitating high-quality collaborative training. Another groundbreaking model, MedGPT \cite{kraljevic2021medgpt}, based on the GPT architecture, utilizes electronic health records to predict future medical events, offering the potential to detect early signs of critical illnesses, such as cancer or cardiovascular diseases, before they are typically diagnosable through conventional methods. Importantly, the utilization of federated learning ensures that sensitive patient data is processed on-site, never leaving the institution’s local environment, thus significantly enhancing data security and maintaining strict patient confidentiality.

The integration of federated learning (FL) with foundation models (FMs) offers unprecedented potential to transform medical diagnostics and personalize treatments, greatly enhancing the capabilities of healthcare systems to deliver exceptional care while adhering to rigorous standards of data privacy and security. This technological advancement not only improves patient outcomes but also strengthens trust in the use of AI within critical sectors such as healthcare. However, deploying federated FMs in the biomedical domain comes with significant challenges, including ensuring data privacy and security, achieving model generalization across diverse datasets, and maintaining bias and fairness. Addressing these issues is essential for harnessing the full capabilities of FL FMs in healthcare and biomedical research.

Furthermore, this paper explores future directions and ongoing challenges in the field, emphasizing the importance of real-time learning and adaptation, fostering collaborative innovation, and the generation of synthetic data for both academic and industrial applications within FL frameworks. By overcoming these challenges, researchers and practitioners can fully realize the potential of federated foundation models, leading to revolutionary advancements in healthcare. These efforts will not only contribute to scientific progress but also to the practical, ethical, and efficient implementation of AI technologies in sensitive environments, ultimately benefiting global health outcomes.

\begin{itemize}
    \item We provide a comprehensive review of existing literature on Federated Learning (FL) and Foundation Models (FM) within the biomedical and healthcare domains. This review meticulously categorizes and discusses various aspects such as biomedical and healthcare data sources, foundation models, federated privacy, and downstream tasks, offering a thorough synthesis of current knowledge and methodologies.
    \item We introduce a taxonomy of biomedical healthcare foundation models, classifying the existing representative FMs from diverse perspectives including model architecture, training strategy, and intended application purposes. This taxonomy aids in the systematic understanding and comparison of different models.
    \item We explore the open challenges and outline future research directions for the integration of FL with FMs in the biomedical and healthcare sectors, providing insights into unresolved issues and potential advancements.
    \item To the best of our knowledge, this is the first survey paper to extensively cover foundation models in federated learning specifically tailored for biomedical and healthcare applications. Our survey uniquely addresses both large-language and vision-language models, highlighting their relevance and transformative potential in this context.
\end{itemize}

\paragraph{How do we collect papers?} In this survey, we collect over two hundred related papers in the field of Federated Learning, Foundation Model, and Biomedical healthcare. We consider Google Scholar as our main literature search engine, where the MedPub, Web of Science, and IEEE Xplore are also used as essential tools. Moreover, we check most of the related top-tier conferences, such as NeurIPS, ICML, ICLR, CVPR, and ECCV, and Bioinformatics. The major keywords we use are ``Biomedical Federated Learning, Medical Pretrained Foundation Model, Healthcare Federated Pretrain Training, etc''. The most representative papers like Med-BERT \cite{rasmy2021med}, FedClip \cite{lu2023fedclip}, and MedClip \cite{medclip2022} are regarded as seed papers for reference check.

\paragraph{Organization} The rest of this survey is organized as follows. Sec.~\ref{sec:background} describes the FM and FL literature relevant to our work. Sec.~\ref{sec:flfm-scracth} detail how to apply FMs with FL. The applications of FM on biomedical and healthcare is summarized in Sec.~\ref{sec:fm-medical}. The challenges and future directions of Federated FMs in the biomedical and healthcare sectors are discussed in Sec.~\ref{sec:challenges}. Finally, we conclude our survey in Sec.~\ref{sec:conclusion}.


\section{Background}

\subsection{Background on Foundation Models}\label{sec:background}
The latest wave of AI innovation sees the evolution of a new class of AI models often referred to as foundation models (FMs) - a term popularized by the Stanford Institute for Human-Centered AI \cite{bommasani2021opportunities} which can be categorized into two model types: Large-language Model (LLM) and Vision-language Model (VLM). For example, LLMs like the recent examples from OpenAI including ChatGpt and Gpt-4 \cite{achiam2023gpt} demonstrate impressive capabilities to generate coherent text. While VLMs such as DALL·E 2 \cite{ramesh2022hierarchical} shows the ability to create realistic images and art from a text description. These models are trained with pretraining, self-supervised training, and reinforcement-instructed fine-tuning with broad data at immense scale and high resource costs, resulting in models with billions of parameters \cite{bommasani2021opportunities}. In this section, we will introduce the backbone of Foundation Models in Sec.~\ref{subsec:backbone}, where the pre-trained large-language models and vision-language models are discussed in Sec.~\ref{subsec:llm} and Sec.~\ref{subsec:vlm}, respectively.

\subsubsection{Backbone Networks in Foundation Models}\label{subsec:backbone}

The significant advancements in foundation models are largely due to the evolution of their underlying architectures, transitioning from Long Short-Term Memory networks (LSTM) \cite{hochreiter1997long} to Transformers \cite{vaswani2017attention}. Initially, LSTMs served as the basic architecture for early pre-trained models, where the recurrent structure is computationally intensive when scaled to deeper layers. In response to these limitations, the Transformer architecture was developed and quickly established itself as the standard for modern natural language processing (NLP) \cite{chowdhary2020natural}. The superiority of Transformers over LSTMs can be attributed to two key factors: (1) Efficiency: Transformers eliminate recurrence, enabling parallel computation of tokens.(2) Effectiveness: The attention mechanism facilitates dynamic spatial interactions between tokens, contingent on the input itself. This section provides a brief overview of the evolution of backbone networks in foundation models, highlighting the transition from LSTMs to Transformers, followed by vision language model backbones from Convolutional neural networks (CNNs) \cite{lecun1998gradient} to Vision Transformers (ViTs) \cite{dosovitskiy2020image}. 

\paragraph{Backbone Networks in Texts.}
Transformer has become the backbone of most pre-trained language models, such as BERT \cite{kenton2019bert}, GPT\cite{jeblick2023chatgpt}, and T5\cite{raffel2020exploring}, building upon self-attention module and feed-forward networks (FFNs). The self-attention module facilitates token interaction, while FFN refines token representations using non-linear transformations. The Transformer architecture is designed to process tokens efficiently in parallel, thanks to the elimination of recurrent units and the use of position embeddings. Additionally, the architecture includes residual connections, layer normalization, and other features that prevent saturation issues and enhance expressive power with large-scale data and deep layers. The input is linearly transformed into query, key, value ($Q, K, V$), and output spaces in the self-attention module: the attention scores between the query and key is computed, which are then used to weight the values. The FFN module processes the weighted values to generate the output. The Transformer architecture has proven to be superior in terms of capacity and scalability, enabling the development of increasingly sophisticated language models. Considering an input $X$m the linear transformation of $X$ into $Q, K, V$ is computed as follows:
\begin{equation}\label{eq:transformer}
    Q = XW^Q, K = XW^K, V = XW^V,
\end{equation}
where the self-attention module is calculated with a softmax function as follows:
\begin{equation}\label{eq:selfatt}
    \text{Attention} (Q,K,V) = \text{softmax}(\frac{QK}{\sqrt{d_k}})V.
\end{equation}
To this end, the FFN provides the non-linear features for the transformer architecture. Besides the self-attention and FFN modules, the transformer architecture also includes residual connections \cite{he2016deep}, layer normalization \cite{ba2016layer}, and positional encoding \cite{shiv2019novel} to enhance the model's performance. The transformer architecture has been widely adopted in various pre-trained language models, such as BERT, GPT, and T5, and has been instrumental in advancing the field of natural language processing (NLP).

\paragraph{Backbone Networks in Images.}

Convolutional Neural Networks (CNNs) \cite{lecun1998gradient} have long been the foundation for many vision-related tasks, characterized by their distinctive architecture comprising convolutional, pooling, activation, and fully connected layers. These layers work in unison where convolutional layers act as trainable filters identifying image patterns like edges and textures, pooling layers reduce data dimensionality, activation layers introduce non-linearity, and fully connected layers synthesize these features into predictions. This architecture has not only been pivotal in vision applications but has also been adapted for language understanding tasks.

As the field evolves, there has been a notable shift towards incorporating Transformer architectures, originally designed for natural language processing, into vision tasks. This integration is exemplified by the development of Vision Transformers (ViT)  \cite{dosovitskiy2020image}, which apply the Transformer's self-attention mechanisms to image patches for feature extraction, representing a significant evolution from traditional CNN approaches. This concept has similarly influenced computational biology, as seen in models like AlphaFold2 \cite{chen2015convolutional}, which leverages Transformer technology for protein structure prediction. These adaptations underscore the versatility and robustness of Transformer models across different scientific domains.

\subsubsection{Background on Large Language Models}\label{subsec:llm}
In the field of Natural Language Processing (NLP), the evolution of methods to build token representations has been marked by significant advancements. Initially, typical approaches such as those proposed by \cite{mikolov2013efficient,pennington2014glove} focused on creating 'static word embeddings,' where a one-to-one mapping between words and their vector representations is established. These embeddings are termed 'static' because they do not account for the context in which a word is used, thus limiting their ability to reflect the diverse meanings words can have in different settings.

Recognizing the limitations of static embeddings, there has been a shift towards developing 'contextualized word embeddings.' These representations are dynamic, with the vector for a word varying according to its contextual usage. For instance, the word 'bank' would have different embeddings in 'river bank' compared to 'money bank.' This approach, exemplified by models like ELMo \cite{peters2018deep}, GPT \cite{radford2019language}, and BERT \cite{kenton2019bert}, significantly enhances the quality of word representations by modeling bi-directional contexts, thereby improving performance across various NLP tasks.

Historically, neural language models \cite{bengio2000neural,mikolov2013distributed} served as foundational frameworks in NLP, utilizing relatively shallow neural architectures for efficient training. These models were primarily pre-trained on tasks like unidirectional language modeling, which involves predicting the next word based on previous words. However, subsequent innovations such as Skip-Gram \cite{mikolov2013efficient} aimed to enrich word embeddings by predicting surrounding words or using bidirectional context, respectively. GloVe \cite{pennington2014glove} extended this by focusing on word co-occurrence probabilities.

The advent of deep learning brought about more sophisticated approaches for learning word representations. ELMo \cite{peters2018deep} introduced a bidirectional language modeling task, utilizing both forward and backward context in its pre-training. GPT \cite{radford2019language} continued with unidirectional modeling, while BERT \cite{kenton2019bert} innovated with the Masked Language Model. This method involves masking words in a sentence and predicting them based on the remaining unmasked context, allowing for deeper bidirectional context modeling.

Further developments like the T5 model \cite{raffel2020exploring} introduced an encoder-decoder framework for generating text outputs, proving particularly effective in text generation tasks such as summarization and question-answering. These advancements have been integral to the development of versatile language models like OpenAI's GPT-3, InstructGPT, Codex, and ChatGPT, which not only generate text but can also engage in conversational exchanges, admit errors, and handle complex user interactions.

\paragraph{Representative Large Language Models}

Large Language Models (LLMs) have become pivotal in the evolution of natural language understanding and generation. The progression of the GPT series, from GPT-1 \cite{radford2019language} to GPT-3 \cite{brown2020language}, and the subsequent release of GPT-4 \cite{achiam2023gpt}, illustrates a remarkable expansion in model size and versatility. These models have profoundly impacted AI research and applications, heralding a new era of computational linguistics. Concurrently, BERT \cite{kenton2019bert} revolutionized pre-training approaches by emphasizing bidirectional training, which significantly enhances language understanding capabilities. PaLM \cite{chowdhery2023palm}, another notable advancement, has achieved state-of-the-art results across diverse language tasks, highlighting the potential for scalability in LLMs. Recent innovations also include Bard \cite{manyika2023overview}, which integrates extensive world knowledge into a context-aware framework, and LLaMa \cite{touvron2023llama}, which prioritizes efficiency and practical applicability in language model design. Collectively, these models mark crucial developments in the field, each contributing distinctively to the enrichment and complexity of machine learning techniques that underpin contemporary AI systems.

\subsubsection{Background on Vision Language Models}\label{subsec:vlm}
Deep neural networks have exhibited remarkable success across a variety of vision tasks, such as image classification, object detection, and instance segmentation, largely attributable to the effectiveness of pre-training. Initially, pre-training in the vision domain involved training models on extensive annotated image datasets like ImageNet \cite{deng2009imagenet}. However, to address issues such as generalization errors and spurious correlations inherent in supervised learning, various self-supervised learning methods have been developed.

A significant area of advancement in AI research is the integration of vision and language models, which aims to develop systems capable of understanding and generating content that spans visual and textual modalities. The introduction of the Vision Transformer (ViT) \cite{dosovitskiy2021an} marked a pivotal shift by applying the transformer architecture—originally designed for natural language processing—directly to sequences of image patches. This approach fundamentally changed the paradigm of how models process visual information. Building on this, CLIP (Contrastive Language–Image Pre-training) \cite{radford2021learning} advanced the field by learning visual concepts through natural language supervision, enabling the model to adeptly handle various vision tasks with minimal task-specific training. Further extending these innovations, Stable Diffusion \cite{rombach2022high} ventured into generative art, providing tools to create intricate images from textual descriptions. The most recent breakthrough, Segment Anything \cite{kirillov2023segment}, tackles the complex challenges of image segmentation, using deep learning to precisely identify and delineate multiple objects within images in ways that are contextually relevant. Collectively, these developments not only bridge the gap between visual data and language processing but also set the stage for more intuitive and interactive AI systems.

\subsubsection{Challenges of Foundation Models}
The paradigm of foundation models (FMs) represents a significant shift from traditional task-specific models that have long dominated the AI landscape. These pre-trained models are designed for adaptation to a variety of tasks they were not originally trained for \cite{wojcik2022foundation}. Adaptation techniques include user or engineer prompts, continual learning, and fine-tuning—methods that expand their application to fields where data scarcity impedes the development of specialized algorithms. This flexibility introduces exciting possibilities for scalable, reusable AI models across diverse domains, including transformative potential in healthcare \cite{yu2018artificial}. However, this shift also presents unique challenges, including the risk of over-generalization, the difficulty in fine-tuning for highly specialized tasks, and the ethical implications of deploying such versatile technologies in sensitive areas. These challenges necessitate rigorous validation, careful implementation, and ongoing monitoring to ensure that the deployment of foundation models aligns with ethical standards and practical requirements.

\paragraph{Over-trusting High Performance $\&$ Output Coherence: Ensuring Safe $\&$ Reliable Use}
Despite the high accuracy and broad capabilities of larger models, it's critical to address ethical and legal standards to ensure their use remains safe, fair, and privacy-conscious \cite{wojcik2022foundation}. In healthcare, the necessity for accurate and reliable data for clinical decision-making cannot be overstated. However, verifying the correctness of outputs from FMs poses a challenge, as demonstrated by systems like ChatGPT, whose outputs can mimic human-like text, potentially leading to automation bias and misuse \cite{bender2021dangers}. The complexity of these models often precludes a full understanding of their mechanisms, necessitating cautious deployment decisions, especially in sensitive fields like healthcare. This includes designing interfaces that clearly articulate the limitations and probabilistic nature of AI outputs and developing robust validation processes to ensure safety and fairness.

\paragraph{Building AI in a Vacuum: Decontextualized $\&$ Centralized}
AI development frequently takes place in isolation, focused on technological accuracy before considering real-world user needs \cite{osman2021realizing}. This 'development in a vacuum' has drawn increasing scrutiny for failing to address the actual conditions and requirements of end-users \cite{liao2022connecting}. Foundation models, in particular, suffer from this issue as they require significant adaptation to be truly effective outside of initial testing environments. A greater emphasis on ethnographic studies could provide deeper insights into the practical applications and challenges of AI within operational settings. Moreover, integrating AI technology into everyday use demands an understanding of specific user contexts, necessitating strategies for risk mitigation and a move towards more user-centered research directions. Validating the utility of AI in real-world settings and ensuring their integration into clinical practice remains a formidable challenge, but one that the human-computer interaction (HCI) community is well-equipped to tackle by bridging the 'last mile' of AI in healthcare \cite{zajkac2023clinician}.

\subsection{Background of Federated Learning in Foundation Models}

\subsubsection{Background of Conventional Federated Learning and Frameworks}
Federated Learning (FL) is a machine learning paradigm where multiple clients, such as mobile devices or entire organizations, collaboratively train a model under the orchestration of a central server, such as a service provider, while keeping the training data decentralized. This method not only adheres to the principles of focused collection and data minimization but also addresses many systemic privacy risks and costs associated with traditional centralized machine learning approaches. The concept of FL, first introduced by McMahan et al. in 2016 \cite{mcmahan2017communication}, has grown significantly in interest from both theoretical and practical perspectives. This approach is defined by challenges including unbalanced and non-IID data across numerous unreliable devices, limited communication bandwidth, and the complexities of model training and implementation across diverse and distributed environments.

Since its inception, the focus of federated learning has expanded beyond mobile and edge devices to include applications involving a smaller number of more reliable entities, such as multiple organizations collaborating to train a model. This has led to distinguishing between "cross-device" and "cross-silo" federated learning, each with unique challenges and requirements. In this survey, we delve into the specifics of cross-device federated learning, highlighting its practical aspects, challenges, and its potential to train and implement foundation models (FMs) in a distributed fashion.

Groundbreaking works of FL like that by McMahan et al. \cite{mcmahan2017communication} laid the foundational framework for FL systems. Research in FL has since advanced, focusing on enhancing data privacy in applications such as medical image segmentation \cite{li2019privacy} and addressing ongoing challenges related to communication efficiency, scalability, and model robustness \cite{kairouz2021advances, li2020federated}. Notable developments in FL include algorithms like SCAFFOLD \cite{karimireddy2020scaffold} and FedProx \cite{li2020federated}, which tackle issues such as client update variance and client drift in non-IID data environments. Further contributions from FedGSam \cite{qu2022generalized}, FedLGA \cite{li2023fedlga}, and LoMar \cite{li2021lomar} have advanced FL by developing generalized strategies and adaptive algorithms that enhance learning processes in federated settings.

Moreover, the development of open-source FL frameworks such as TsingTao \cite{Zhang2023fedcp}, Flower \cite{beutel2020flower}, FedML \cite{he2020fedml}, FATE \cite{liu2021fate}, and FederatedScope \cite{xie2023federatedscope} has significantly advanced the accessibility and standardization of FL practices. Designing specialized FL systems and benchmarks is imperative to meet the unique needs and challenges of foundation models (FM). Although current FL frameworks have made significant strides in both academic and industrial settings \cite{bonawitz2019,he2020fedml, huba2022papaya,beutel2020flower,liu2021fate, xie2023federatedscope}, they may not fully satisfy the specific requirements for optimizing memory, communication, and computational demands associated with FMs. Platforms like FedML \cite{he2020fedml} and FATE \cite{liu2021fate} are beginning to adapt to better support FMs, but extensive research is still needed to thoroughly explore system requirements and integration strategies for these models.

\subsubsection{Motivations of Federated Learning for Foundation Models}

\paragraph{Scarcity of Compliant Large-Scale Data}
The shortage of large-scale, high-quality, legally compliant data has become a critical driver for the adoption of federated learning in the context of foundation models. This scarcity is particularly acute in sectors such as technology and social media, where data compliance and privacy issues are increasingly foregrounded \cite{businessinsiderElonMusks, cnnRedditSparks, wiredStackOverflow}.

\paragraph{High Computational Resource Demand}
Training large-scale foundation models demands significant computational resources. For instance, training LLaMa with 65 billion parameters required 2048 NVIDIA A100 GPUs over 21 days \cite{touvron2023llama}, while the smaller 1.3 billion parameter GPT-3 model needed 64 Tesla V100 GPUs for a week \cite{brown2020language}. The development of GPT-4 also highlighted these intensive demands, utilizing substantial resources over several months at considerable financial costs \cite{achiam2023gpt}. Federated learning can help alleviate these demands by distributing computational tasks across multiple devices, thereby optimizing resource utilization.

\paragraph{Continuous Model Updating Challenges}
As data continually evolves, particularly from sources like IoT sensors and edge devices, keeping foundation models updated becomes a significant challenge \cite{hadsell2020embracing, li2024adaer}. Federated learning offers a dynamic solution by enabling ongoing, incremental updates to FMs with new data, which allows these models to adapt to emerging data landscapes without the need to reinitiate training processes. This approach not only enhances the models' accuracy and relevance but also ensures their adaptability to real-world changes \cite{yoon2021federated}.

\paragraph{Reducing Response Delays and Enhancing FM Services}
One of the foremost benefits of applying federated learning to foundation models is the potential to deliver nearly instant responses, thus significantly improving user experience. Traditional central server deployments often face latency and privacy issues due to the required network communications between users and servers \cite{li2024snapfusion}. Federated learning addresses these concerns by enabling models to operate directly on local devices, minimizing network dependencies, reducing latency, and improving privacy protections. This approach not only enhances response times but also ensures a seamless, privacy-conscious interaction, maintaining user trust and satisfaction in the services provided by foundation models.

\subsubsection{Motivations of Foundation Models for Federated Learning}
Foundation Models can significantly contribute to enhancing the efficacy of Federated Learning. This section explores the motivations behind leveraging FM within FL, examines the challenges posed by this integration, and discusses the potential opportunities it offers to the field.

\paragraph{Data Privacy and Shortage Dilemma in FL}
In federated settings, clients often grapple with limited or imbalanced datasets, especially in federated few-shot learning contexts \cite{shome2021fedaffect}. Such data scarcity can result in suboptimal model performance, as it may not fully capture the diversity of the data distribution \cite{gao2022survey}. Moreover, privacy concerns are intensified due to the potential for sensitive information recovery from model updates in FL \cite{lyu2020towards,chen2022practical}. These issues are particularly acute in sectors like healthcare or finance, where data privacy regulations or the inherent sensitivity of the data restrict availability, thus complicating the training process and limiting FL’s effectiveness in these crucial areas. One promising solution is the use of synthetic data generated by FMs. Being extensively pre-trained on vast datasets and further refined through techniques such as fine-tuning and prompt engineering, FMs possess a deep understanding of complex data distributions, enabling them to produce synthetic data that closely mirrors real-world diversity.

\paragraph{Performance Dilemma in FL}
FL can mitigate issues related to non-IID and biased data by leveraging the advanced capabilities of FMs, thus enhancing performance across various tasks and domains \cite{yu2023federated}. FMs can improve FL’s efficiency in several ways. (1). Starting Point Advantage: FMs provide a robust starting point for FL. Clients can begin fine-tuning directly on their local data instead of starting from scratch, leading to faster convergence and enhanced performance while reducing the need for extensive communication rounds \cite{chen2022importance,tan2022federated}. (2). Data Diversity Enhancement: FMs act as powerful generators that can synthesize diverse data, enriching the training dataset in FL. An example is GPT-FL \cite{zhang2023gpt}, which utilizes generative models to produce synthetic data that improves downstream model training on servers. This approach not only boosts test accuracy but also enhances communication and client sampling efficiency. (3). Knowledge Distillation: FMs can address performance issues in FL by acting as knowledgeable teachers through techniques like knowledge distillation \cite{li2019fedmd}.

\paragraph{New Sharing Paradigm Empowered by FM}
Unlike traditional FL, which involves sharing high-dimensional model parameters, FMs facilitate a new paradigm through prompt tuning. PROMPTFL \cite{guo2023promptfl} showcases how FM capabilities can be leveraged to efficiently combine global aggregation with local training on sparse data. This approach focuses on training prompts rather than the entire model, thereby optimizing resource use and enhancing performance. Building on this concept, FedPrompt \cite{zhao2022reduce} introduces an innovative prompt tuning method specifically designed for FL, while a recent study FedTPG \cite{qiu2024federated} explores a scalable prompt generation network that learns across multiple clients, aiming to generalize to unseen classes effectively.

\subsection{Machine Learning/Artificial Intelligence in Biomedical and Health Care}

\subsubsection{AI in Biomedical Healthcare: Data Fusion}
Data is the cornerstone of sense-making in artificial intelligence (AI), playing a crucial role in various sectors, including healthcare, where it arises from diverse sources like care providers, insurers, and academic publications \cite{crawford2021atlas, wojcik2022foundation}. It varies in form (e.g., clinical notes, medical images), scale (e.g., patient versus population level), and style (professional versus lay language), posing both unique opportunities and challenges for the application and training of AI models. Despite the proficiency of machine learning methods in managing and extracting insights from vast, multi-dimensional data \cite{thieme2023designing}, it is vital to address how societal biases and inequalities are embedded in the data. Disparities can manifest in various aspects of healthcare, such as the prioritization of certain medical issues and the exclusion or misrepresentation of specific population groups. These issues often stem from barriers like limited healthcare access, restrictive criteria for clinical trial participation, or the risk of inaccurate data due to documentation errors and systemic discrimination \cite{chen2021ethical}. For example, in California, the mandate to verify citizenship at hospitals has reduced autism diagnosis rates among Hispanic children in the context of stringent federal immigration policies.

Healthcare and biomedicine are major sectors within the U.S. economy, accounting for about 17\% of the Gross Domestic Product (GDP) \cite{swensen2011controlling,van2019artificial, keehan2020national}. These fields require substantial financial investments and extensive medical knowledge, encompassing everything from patient care to the scientific exploration of diseases and the development of new therapies \cite{yu2018artificial,korngiebel2021considering}. We envision machine learning models as central repositories of medical knowledge, trained on a diverse array of data sources and modalities within medicine \cite{krumholz2016data,suresh2020deep}. These models could serve as dynamic platforms that medical professionals and researchers use to access and contribute to the latest findings, enhancing their ability to make informed decisions \cite{ionescu2020deep}.

\paragraph{Biomedical Data Fusion}
In the field of biomedical research, a significant challenge lies in deciphering the complex interactions within and between the cellular and organismal levels, characterized by diverse components that exhibit emergent behaviors \cite{ma2017complex}. The data collected through various sensors, while rich, often provides limited insights when examined in isolation due to the specificity of each measurement modality \cite{ramachandram2017deep}. Data fusion, the process of integrating data from multiple modalities, aims to provide a holistic view of biological phenomena by combining disparate data sources that offer unique perspectives on the same subject \cite{hall1997introduction}. This approach is generally advantageous in several ways, categorized into complementary, redundant, and cooperative features of the data \cite{castanedo2013review}. These features are not mutually exclusive but interact synergistically, enhancing the robustness and accuracy of the insights gained.

The goal of effective data fusion strategies is to optimally exploit these complementary, redundant, and cooperative aspects. This requires the use of sophisticated machine learning (ML) methods capable of integrating both structured and unstructured data while accommodating their varied statistical properties, sources of non-biological variation, high dimensionality, and distinct patterns of missing values \cite{li2018review, ramachandram2017deep}. A comprehensive examination of these strategies is presented in a review of multimodal deep learning applications in biomedical data fusion, which outlines the potential advancements and methodologies in the field \cite{stahlschmidt2022multimodal}.

\paragraph{Categories Summary of Data Fusion}

The categories of data fusion techniques can be broadly summarized into three main approaches: easy fusion, intermediate fusion, and late fusion. Easy fusion typically involves direct modeling techniques where different types of neural networks are used to process the input data. This includes fully connected networks for a straightforward integration of features across modalities \cite{park2020prediction}, convolutional networks that are effective in handling spatial data \cite{peng2019capsule}, and recurrent networks suited for sequential data integration \cite{bichindaritz2021integrative}. Autoencoders also play a significant role in easy fusion, with variations such as regular \cite{chaudhary2018deep}, denoising \cite{franco2021performance}, stacked \cite{islam2020integrative}, and variational autoencoders \cite{albaradei2021metacancer} being employed to refine the fusion process.

Intermediate fusion, on the other hand, involves branching strategies that can be homogeneous, focusing on either marginal \cite{lee2019predicting} or joint representations \cite{suk2014hierarchical}, or heterogeneous, which also targets both marginal \cite{xu2021accurately} and joint data representations \cite{wang2021deep}. These strategies optimize the integration by selectively focusing on how data from the same or different modalities are fused.

Late fusion utilizes aggregation methods to combine features at a higher level, often after initial independent processing. Techniques in this category include simple averaging \cite{soto2022multimodal} and weighted averaging \cite{sun2018multimodal}, where weights might be assigned based on the reliability or importance of each modality. Furthermore, meta-learning approaches are utilized to dynamically adjust these weights for optimal performance \cite{huang2020multimodal}, thus enhancing the fusion's effectiveness by incorporating learning-based adjustments. These methods ensure that the final model output maximally benefits from the diverse characteristics of all data modalities involved.

\subsubsection{Conventional AI in Biomedical Healthcare: Segmentation}
\paragraph{Medical Image Segmentation Application}
Medical image segmentation is an essential technique in clinical studies, significantly aiding in disease progression monitoring and diagnosis \cite{worth1997neuroanatomical}. This process involves dividing a medical image into several regions of interest (ROIs), each isolating distinct anatomical structures suitable for various medical applications. Segmentation facilitates a wide range of activities, such as disease diagnosis, pathology localization, tissue volume quantification, treatment planning, computer-assisted surgery, and anatomical studies. Despite manual segmentation being the gold standard for in vivo images, it is labor-intensive, costly, and susceptible to human error due to fatigue, which underscores the need for dependable automated segmentation techniques \cite{mharib2012survey,grimson1997utilizing}.

Medical imaging technologies like magnetic resonance imaging (MRI), computed tomography (CT), and positron emission tomography (PET) typically produce images represented as stacks of 2D slices in a 3D configuration \cite{song2012automated}. The inherent challenges in medical imaging include partial volume effects, noise, insufficient resolution, and the differentiation of complex structures amidst noise and inhomogeneity. The intricate nature of human organs further complicates segmentation tasks, with organs such as the kidneys comprising diverse structures like the renal column, cortex, pelvis, and medulla. Additionally, boundaries between adjacent organs, for example, between the liver and spleen, often blur, making accurate segmentation difficult \cite{hu2009survey}.

In many cases, a 3D medical image is composed of a series of 2D slices, and its dimensions (width, height, depth) are denoted by \(w \times h \times d\) varying according to the specific organ and resolution \cite{taha2015metrics}. The segmentation process involves partitioning these anatomical structures within an image \(I\) into distinct regions, each corresponding to different anatomical features.

Segmentation as an image processing task involves labeling each pixel (in 2D) or voxel (in 3D) according to the class it belongs to, which represents a particular region of interest. Traditional labeling methods include region-based segmentation, watershed and deformable models, and thresholding techniques. The effectiveness of these techniques is significantly influenced by factors like object inhomogeneity, noise, and contrast variation \cite{bajcsy2015survey}. Additionally, segmentation can be approached as an optimization problem where techniques such as energy minimization and maximum a posteriori probability are utilized to achieve the best segmentation outcomes \cite{huang2017breast}. This structured and automated approach aims to overcome the limitations of manual segmentation, offering more consistent and precise results essential for advanced medical applications.

Biomedical image segmentation is a pivotal component of medical imaging, essential for delineating anatomical structures across various imaging modalities. This process is crucial for clinical analysis and medical interventions, enabling precise measurements and enhancing understanding of anatomical features through non-invasive techniques. Recent surveys have extensively examined the application of machine learning methods to biomedical image segmentation, providing detailed insights into both technical approaches and cutting-edge applications within the field. For instance, Seo et al. explore a spectrum of machine learning strategies, ranging from traditional algorithms to sophisticated deep learning models that have substantially enhanced the precision and efficiency of image segmentation \cite{seo2020machine}. Similarly, Alzahrani et al. present an exhaustive survey focused on the methodologies and practical applications of image segmentation in biomedical contexts \cite{alzahrani2021biomedical}. Their work underscores both the advancements and the ongoing challenges in this area of research, emphasizing the dynamic nature of this field and its significant impact on healthcare and medical sciences.

\subsubsection{FM in Biomedical Healthcare}

\paragraph{Motivations}
Foundation models hold transformative potential for biomedical research, particularly in the realms of drug discovery and disease understanding, thereby enhancing healthcare solutions \cite{hanney2015long}. Biomedical discovery processes are currently characterized by intensive demands on human resources, lengthy experimental timelines, and substantial financial outlays. For example, the drug development journey includes stages from basic research, such as protein target identification and potent molecule discovery, through clinical development involving clinical trials, to the final drug approval stage. This extensive process typically spans more than a decade and incurs costs often exceeding one billion dollars \cite{wouters2020estimated}. Thus, the ability to expedite biomedical discovery by harnessing existing data and published findings becomes crucial, especially during critical times like the COVID-19 outbreak, which resulted in significant loss of life and economic damage \cite{lalmuanawma2020applications}.

Foundation models contribute to biomedical advancements in two primary ways. Firstly, these models exhibit strong generative capabilities, as seen with coherent text generation in models such as GPT-3. These capabilities can be utilized in generating experimental protocols for clinical trials and in designing novel molecules for drug discovery \cite{kadurin2017drugan, harrer2019artificial}. Secondly, foundation models excel at integrating diverse data modalities in medicine, facilitating the exploration of biomedical concepts across various scales—from molecular to patient and population levels—and integrating multiple knowledge sources, including imaging, textual, and chemical data \cite{lanckriet2004statistical, aerts2006gene, kong2011integrative, ribeiro2012classification, wu2021babel}. This integrated approach enables discoveries that might be challenging with single-modality data alone.

Additionally, foundation models are adept at transferring knowledge across different data modalities. For instance, research by Lu et al. \cite{lu2022frozen} demonstrated how a transformer model, initially trained on natural language, a data-rich modality, could be adapted for other sequence-based tasks, such as protein folding predictions, a longstanding challenge in biomedicine. These capabilities highlight the potential applications of foundation models in addressing complex biomedical tasks.

\paragraph{Applications}
Foundation models (FMs) hold significant potential for revolutionizing healthcare applications through their adaptability and efficiency in performing specific healthcare and biomedical tasks. They have been proposed for use in a variety of areas including disease prediction \cite{rasmy2021med}, triage or discharge recommendations \cite{korngiebel2021considering}, and health administration tasks such as clinical notes summarization \cite{krishna2021generating} and medical text simplification \cite{jeblick2023chatgpt}. These applications leverage the unique capabilities of FMs, such as fine-tuning and prompting \cite{brown2020language}, to tailor solutions to specific needs, enhancing both the accuracy and efficiency of medical services.

FMs are particularly effective in patient-facing roles, such as question-answering systems and clinical trial matching applications, benefiting both researchers and patients by simplifying access to information and streamlining patient recruitment processes \cite{klasnja2012healthcare,zhu2019hierarchical, harrer2019artificial,beck2020artificial}. As central interfaces, FMs facilitate interactions among data, tasks, and individuals, improving the operational efficiency of healthcare services. This is further explored in subsequent sections focusing on specific healthcare and biomedical tasks.

Additionally, FMs serve as repositories of extensive medical knowledge, accessible by healthcare professionals and the public for purposes like medical question-answering and interactive chatbot applications. Innovations such as ChatGPT \cite{zhang2023gpt} and Bard \cite{pichai2023bard} provide conversational user interfaces that assist users in navigating complex health information and obtaining relevant health advice.

The implementation of FMs also promises to accelerate healthcare application development and research. These models can automate processes such as structured dataset generation, data labeling, and synthetic data creation \cite{chen2021synthetic}. Looking forward, there is considerable scope for developing new FM-enabled capabilities, particularly through the use of multimodal data, a feature characteristic of the healthcare domain. Beyond natural language processing, breakthroughs are already evident in areas like biomedical research, where tools like AlphaFold \cite{tunyasuvunakool2021highly} have made significant advances in predicting human protein structures to aid drug development. Similarly, innovations in genome sequencing are hastening the detection of disease-causing genetic variants \cite{goenka2022accelerated}, and new methods are being developed for optimized clinical trial design \cite{chien2022multi}. This multidisciplinary integration highlights the transformative potential of FMs in enhancing and expanding the capabilities of the healthcare sector.


\section{Federated Learning and Foundation Models}\label{sec:flfm-scracth}

Federated learning (FL) and foundation models represent two cutting-edge approaches in the field of machine learning. Federated learning offers a decentralized approach to training models across multiple nodes or devices, ensuring privacy and maintaining data locality. In contrast, foundation models, due to their vast size and generalized pre-training, provide significant adaptability and scalability for a variety of tasks. Integrating these technologies poses unique challenges but also opens up exciting opportunities for innovation in AI training and application. This section explores the relationship between federated learning and foundation models, highlighting key research directions and recent advancements in this domain. Depending on the training paradigm, foundation models can either be trained from scratch or fine-tuned on top of pre-trained models within a federated learning framework. Additionally, the application of foundation models in federated learning can extend to large language models or vision language models.

Related surveys further enrich our understanding of this integration. A recent survey by Yu et al. \cite{yu2023federated} discusses the intersection of foundation models with federated learning, exploring the motivations behind their integration, the challenges faced, and future directions for research in this domain. This survey serves as a crucial resource for comprehending the current landscape and the potential of combining federated learning strategies with the robust capabilities of foundation models. Another pivotal work by Zhuang et al. \cite{zhuang2023foundation} provides an in-depth analysis of how foundational models can be effectively adapted and optimized within a federated learning framework, discussing both the technical hurdles and the potential breakthroughs. Additionally, Kairouz et al. \cite{kairouz2021advances} offers a comprehensive overview of the advancements and persistent challenges in federated learning, highlighting issues such as algorithm efficiency, data heterogeneity, and security concerns. These surveys collectively offer a rich tapestry of insights into the evolving field of federated learning and foundation models, emphasizing their complexities and transformative potential.

\subsection{Federated Learning and Foundation Models}

The integration of pre-training techniques within federated learning (FL) setups, especially for large-scale models, is increasingly viewed as essential for boosting model performance and broadening their applicability. Extensive research underscores the importance of pre-training in preparing large models to face the unique challenges presented by the decentralized nature of federated datasets. Chen et al. highlight the vital role of pre-training in readying large models for these challenges, emphasizing its necessity for effective performance within federated learning frameworks \cite{chen2022importance}. In a similar vein, Nguyen et al. investigate how the initial conditions of model training, such as the starting points of pre-training and model initialization, critically influence the effectiveness and convergence of FL models \cite{nguyen2022begin}. These studies stress the importance of meticulous pre-training phases to ensure that large models are fully equipped to navigate the complexities of federated learning, thereby maximizing their performance and utility in diverse applications. The application of federated learning to foundation models not only addresses these preparatory needs but also leverages the inherent strengths of both paradigms to offer several key advantages: (1). Efficient Distributed Learning: Federated learning enables models to learn from data distributed across multiple devices or servers without needing to centralize the data, thus preserving privacy and reducing data movement costs. (2). Parameter-efficient Training: By utilizing techniques such as model compression and prompt tuning within a federated framework, the training process becomes more parameter-efficient. This is particularly beneficial in environments where computational resources are limited. (3). Prompt Tuning: This method involves fine-tuning a model on a specific task by adjusting a small set of parameters, and when combined with federated learning, it allows for personalized model tuning on decentralized data.
(4). Model Compression: Techniques like quantization and pruning that reduce the model size can be effectively applied in federated settings, enhancing the feasibility of deploying large models on edge devices with limited storage and processing capabilities.

\begin{figure}[tb]
	\centering
	\begin{subfigure}{0.32\columnwidth}
		\includegraphics[width = 1\columnwidth]{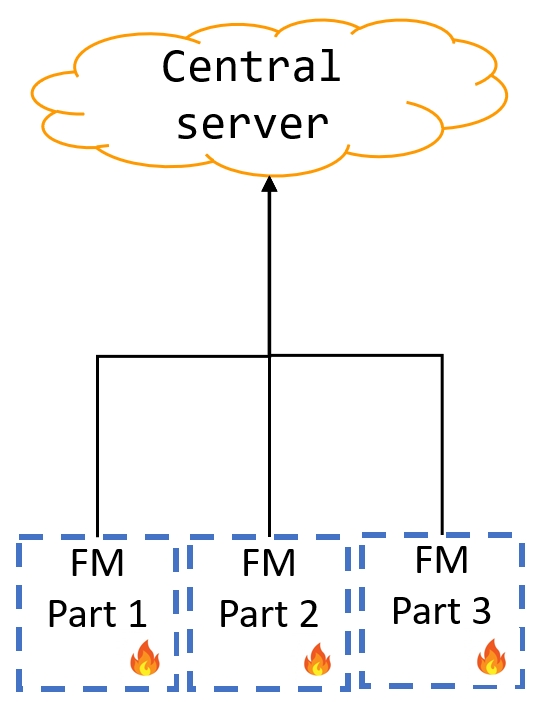}
		\caption{FLFM: Model parallelism.}
		\label{fig:flfm-model}
	\end{subfigure}
	\begin{subfigure}{0.32\columnwidth}
		\includegraphics[width = 1\columnwidth]{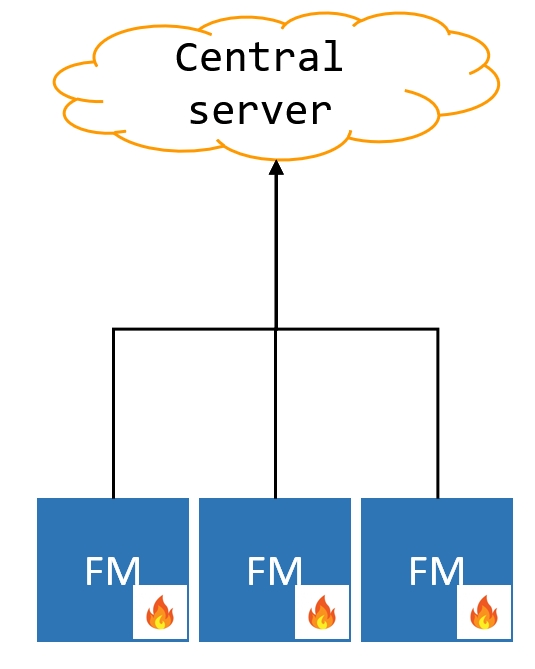}
		\caption{FLFM: Pipeline parallelism.}
		\label{fig:flfm-pipeline}
	\end{subfigure}
	\begin{subfigure}{0.32\columnwidth}
		\includegraphics[width = 1\columnwidth]{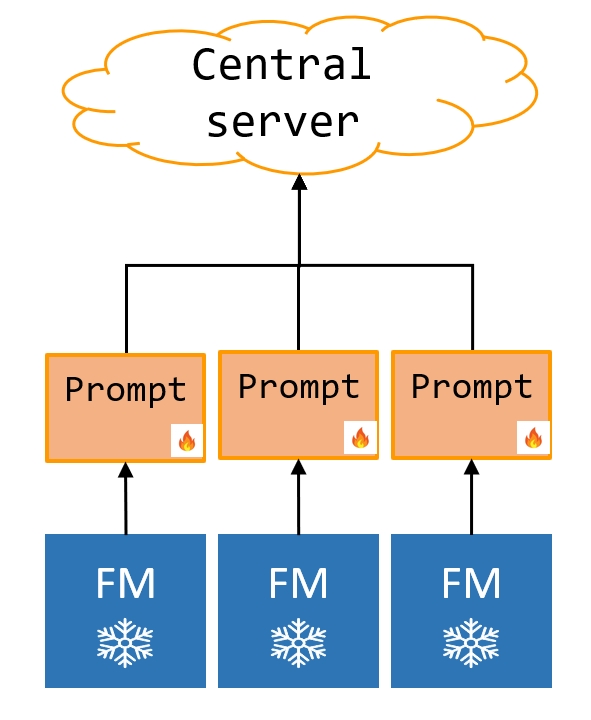}
		\caption{FLFM: Parameter efficient.}
		\label{fig:flfm-prompt}
	\end{subfigure}
	\caption{Efficient distributed learning and parameter efficient strategies for foundation models in federated learning.}
	\label{fig:flfm_efficient}
\end{figure}

\paragraph{Efficient Distributed Learning Algorithms}
Efficient distributed learning algorithms are critical for optimizing foundation models within the constraints of limited resources \cite{zhao2022reduce,wu2022communication}. These algorithms are specifically engineered to address the twin challenges of enhancing communication and computation efficiency during the training and deployment of large FMs across a network of devices, which may vary in capabilities and network conditions. Two pivotal techniques in this regard are model parallelism and pipeline parallelism.

Model parallelism \cite{liao2023accelerating} involves dividing the model into different segments and distributing these segments across multiple devices. This allows for simultaneous processing and can significantly expedite the computation process by leveraging the combined power of multiple devices. On the other hand, pipeline parallelism \cite{wang2022pipefl} focuses on enhancing the overall system's efficiency and scalability by organizing the computation process in stages. Each stage can be processed on different devices in a pipeline manner, thus optimizing the workflow and reducing idle times.

An illustrative example of these parallelism strategies in federated learning (FL) for FMs is demonstrated in Figure.~\ref{fig:flfm_efficient}, where participants train distinct layers of a model using their own private, local data. Note that Fig.~\ref{fig:flfm-model} is the illustration of model parallelism and Fig.~\ref{fig:flfm-pipeline} demonstrates the pipeline parallelism. This approach not only maintains the privacy of the data but also contributes to the efficiency of the learning process. Recent studies, such as the research conducted by Yuan et al., validate the practicality of utilizing pipeline parallelism for decentralized FM training across heterogeneous devices \cite{yuan2022decentralized}.

\paragraph{Parameter-efficient Training Methods}

Parameter-efficient training methods are increasingly critical in optimizing foundation models for specific domains or tasks. These methods typically involve integrating adapters—a technique where the core parameters of the FM are frozen, and only a small, task-specific section of the model is fine-tuned. This approach is illustrated in Figure~\ref{fig:flfm-prompt}, which shows how adapters can be effectively incorporated into the federated learning framework for FMs. Recent implementations such as FedCLIP \cite{radford2021learning} and FFM \cite{yu2023federated} utilize this method to fine-tune FMs, achieving substantial performance improvements.

By focusing adjustments on small adapters rather than the entire model, these training methods greatly reduce the computational and communication demands typically required \cite{houlsby2019parameter,hu2022lora}. This is particularly beneficial in FL environments where conserving bandwidth and processing power is crucial due to the distributed nature of the data and the varying capacities of participating devices. However, despite these efficiencies, the underlying requirement for substantial computational resources to manage the FM and execute the fine-tuning process remains significant.

\paragraph{Prompt Tuning}
Prompt tuning has rapidly gained traction as a communication-efficient alternative to full model tuning, demonstrating effectiveness comparable to more resource-intensive methods \cite{lester2021power}. This technique involves fine-tuning lightweight, additional tokens while keeping the foundational model's main parameters frozen, which avoids the necessity of sharing large model parameters across the network. In federated learning scenarios, this approach enables leveraging the collective knowledge from multiple participants to refine the prompts used in FM training, potentially enhancing the performance of the FM.

The integration of prompt tuning in FL, similar to the parameter efficient approach depicted in Figure~\ref{fig:flfm-prompt}, has been explored in recent research. Studies such as FedPrompt \cite{zhao2023fedprompt} and PROMPTFL \cite{guo2023promptfl} have shown promising results by improving the quality and effectiveness of prompt-based training methods through FL frameworks. These methods enable efficient and targeted tuning of model behaviors without requiring extensive data transfer or the deployment of large-scale models on each participant's device, thereby conserving bandwidth and computational resources.

Moreover, a recent study, FedTPG \cite{qiu2024federated}, investigates a scalable prompt generation network that learns across multiple clients, aiming to generalize effectively to unseen classes. This approach demonstrates the potential of FL to enhance the sophistication of prompt tuning methodologies by distributing the learning process across a wide array of devices and data sources.

However, the implementation of prompt tuning in FL is not without challenges. Concerns include the assumptions that large FMs are readily available on user devices, which may not always be feasible in resource-constrained environments. Additionally, there are potential privacy risks associated with utilizing cloud-based FM APIs, which could compromise the security of sensitive data.

\paragraph{Model Compression}
Model compression has emerged as a vital strategy to mitigate the substantial memory, communication, and computational demands of large foundation models. By minimizing the size of these models, model compression enables more practical deployments within federated learning frameworks without significantly compromising performance. Prominent compression techniques include knowledge distillation, where a smaller model is trained to emulate the performance of a larger one \cite{hinton2015distilling}, and quantization, which reduces the numerical precision of model parameters to decrease both size and computational complexity \cite{yang2019quantization}. Additionally, pruning eliminates superfluous or redundant model parameters, significantly lowering the resource requirements of the model \cite{blalock2020state}.  

Implementing these compression techniques effectively requires striking a balance between reducing model size and preserving essential capabilities. This balance ensures that the compressed model performs robustly in real-world applications, maintaining the functionality of the foundation model while reducing operational demands. Therefore, research and development in model compression focus not only on shrinking model dimensions but also on enhancing efficiency and intelligence, tailored for specific deployment scenarios. \cite{song2023resfed} introduces ResFed, a framework that leverages model compression in federated learning to significantly cut down on bandwidth and storage needs while maintaining high model accuracy. \cite{hinton2015distilling} presents the concept of knowledge distillation, which allows a compact "student" model to learn effectively from a larger "teacher" model, thus enabling the student to achieve similar performance with much lower computational costs. \cite{yang2019quantization} explores quantization techniques for training neural networks that perform inference using only integer arithmetic, substantially lightening model load without sacrificing accuracy.  \cite{blalock2020state} provides a thorough review of neural network pruning techniques, showing their potential to significantly reduce model size while maintaining or improving performance. \cite{li2020federated} discusses the integration of model compression into federated learning, tackling challenges related to efficiency and scalability in privacy-preserving, decentralized machine learning.

\subsection{FL on Large Language Models and Vision Language Models}
In this section, we delve into the integration of federated learning with foundation models on the two main model applications: large language models and vision language models, exploring the unique challenges and opportunities presented by these advanced AI systems. 

\subsubsection{FL on Large Language Models}
Federated learning applied to large language models (LLMs) represents a transformative approach to harnessing decentralized datasets for model training, while prioritizing data privacy and security \cite{wu2020fedmed, nagy2023privacy}. This method is especially crucial for LLMs because of their inherent requirement for vast and varied data inputs to accurately capture and interpret the complexities of human language.

This nature of FL effectively addresses privacy concerns by ensuring that sensitive or proprietary data does not leave its original location, thereby reducing the risk of data breaches. Additionally, this decentralized approach allows LLMs to learn from a wider array of linguistic inputs, reflecting regional dialects, colloquialisms, and cultural nuances that might not be present in a centralized dataset \cite{liu2021federated}.

Moreover, the application of FL to LLMs facilitates the development of models that are not only linguistically comprehensive but also more personalized and responsive to local contexts. By training on diverse datasets that are geographically dispersed, LLMs can develop a deeper understanding of language variations and user-specific preferences, leading to improved performance in tasks such as language translation, sentiment analysis, and contextual understanding.

This method also helps in mitigating biases that are often present in centralized training datasets. Since FL involves multiple datasets that are not centrally collected, the resulting model is trained on a broader spectrum of data sources, which can contribute to more balanced and fair outputs. Thus, federated learning not only enhances the privacy and security of data used in training LLMs but also boosts the models' ability to decipher and utilize the full richness of human language, making them more accurate and effective in real-world applications.

\paragraph{Practical Applications of FL on LLMs}
The integration of federated learning with large language models is yielding groundbreaking frameworks and methodologies that significantly enhance language model training while adhering to data privacy and security protocols. In this survey, we highlight several notable applications and advancements in this domain, including:

\begin{itemize}
    \item \textit{Privacy-preserving Federated Learning and its application to natural language processing:} \cite{nagy2023privacy} explores privacy-preserving techniques in federated learning for training large language models. It particularly focuses on models such as BERT and GPT-3, providing insights into how federated learning can be leveraged to maintain privacy without sacrificing the performance of language models in NLP applications.
    \item \textit{FedMed: A federated learning framework for language modeling:} \cite{wu2020fedmed}  introduces ``FedMed", a novel federated learning framework designed specifically for enhancing language modeling. The framework addresses the challenge of performance degradation commonly encountered in federated settings and showcases effective strategies for collaborative training without compromising on model quality.
    \item \textit{Efficient Federated Learning with Pre-Trained Large Language Model Using Several Adapter Mechanisms:} \cite{kim2024efficient} highlights a method to enhance federated learning efficiency by integrating adapter mechanisms into pre-trained large language models. The study emphasizes the benefits of using smaller transformer-based models to alleviate the extensive computational demands typically associated with training large models in a federated setting. The approach not only preserves data privacy but also improves learning efficiency and adaptation to new tasks.
    \item \textit{OpenFedLLM:} This contribution is a seminal effort in federated learning specifically designed for large language models. The "OpenFedLLM" framework facilitates the federated training of language models across diverse and geographically distributed datasets. A standout feature of this framework is its capability to ensure data privacy during collaborative model training. It also incorporates federated value alignment, a novel approach that promotes the alignment of model outputs with human ethical standards, ensuring that the trained models adhere to desirable ethical behaviors \cite{ye2024openfedllm}. Moreover, OpenFedLLM is open-source\footnote{https://github.com/rui-ye/OpenFedLLM} , making it accessible to the broader research community and fostering collaboration in the development of federated language models.
    \item \textit{Pretrained Models for Multilingual Federated Learning:} This study addresses the complex challenges of utilizing pretrained language models within a federated learning context across multiple languages. Weller et al.'s work is crucial for understanding how multilingualism impacts federated learning algorithms, particularly exploring the effects of non-IID (independently and identically distributed) data inherent in natural language processing tasks across different languages. The research explores three main tasks: language modeling, machine translation, and text classification, providing valuable insights into the adaptability of federated learning to diverse linguistic datasets \cite{weller2022pretrained}.
    \item  \textit{GPT-fl:} This innovative approach integrates federated learning with prompt-based techniques to train large language models. "GPT-fl" employs prompt learning within a federated framework, which allows for efficient learning from decentralized data sources while maintaining data privacy. This method enhances model adaptability and performance across various linguistic tasks, making it a promising solution for applications requiring high levels of customization and responsiveness to user-specific contexts \cite{zhang2023gpt}.
\end{itemize}

\subsubsection{FL on Vision Language Models}
The integration of federated learning with vision language models (VLMs) marks a significant advancement in multimodal learning where both visual and textual data are processed in a privacy-preserving, distributed learning environment. These models are crucial for tasks that necessitate a deep understanding and generation of information from visual cues and textual descriptions. Federated learning enhances the capability of VLMs by enabling them to learn from a diverse set of decentralized data sources, including images and associated annotations from various geographic and demographic distributions without the need to centralize sensitive data.

VLMs integrated with FL are particularly beneficial in scenarios where data privacy is paramount, such as in healthcare for patient image data or in surveillance where personal data protection is critical. By processing data locally and only sharing model updates, FL preserves the privacy and security of the underlying data, while still benefiting from the diverse data attributes necessary for robust model training.

This approach also allows for the training of more personalized and region-specific models, capturing a wide array of cultural and contextual nuances in visual-textual datasets. For example, a VLM trained via federated learning can better understand and generate language descriptions for regional landmarks or culturally specific events, enhancing its applicability across different global contexts.

Moreover, the decentralized nature of FL helps in mitigating dataset bias, a common issue in centralized training datasets. Since the training data in FL comes from a wide range of sources, the models are less likely to overfit to the biases present in a single dataset, leading to more generalizable and fair VLMs.

This section underscores the crucial role of prompt learning in expanding the capabilities of both language and vision models trained in federated environments. By facilitating efficient task adaptation and maintaining data privacy, prompt learning represents a significant step forward in the development of AI systems that can operate across diverse and distributed data landscapes.

\paragraph{Practical Applications of FL on VLMs}
\begin{itemize}
    \item \textit{FedCLIP:} Pioneering the field of federated vision-language models, FedCLIP \cite{lu2023fedclip} adapts the powerful CLIP (Contrastive Language-Image Pre-training) architecture \cite{radford2021learning} to operate in a federated setting. Unlike traditional learning models that centralize data, FedCLIP enables collaborative learning across decentralized image datasets with accompanying text descriptions. Crucially, this approach safeguards data privacy by eliminating the need for sensitive user data to leave local devices.
    \item \textit{PromptFL:} \cite{guo2023promptfl} demonstrates the power of combining federated learning with prompt learning techniques for training models on distributed visual and textual data. Prompt learning injects flexibility into model training. In PromptFL, federated learning preserves privacy while prompt learning improves training effectiveness and efficiency across diverse datasets.
    \item \textit{FedPrompt:} Communication-Efficient and Privacy-Preserving Prompt Tuning in Federated Learning \cite{zhao2023fedprompt} addresses two critical aspects of federated learning for vision-language models: efficiency and privacy. Prompt tuning offers adaptability but can be communication-intensive. FedPrompt explores methods to reduce communication overhead while still reaping the benefits of prompt tuning, all while ensuring that sensitive data remains protected.
    \item \textit{pFedPrompt:} \cite{guo2023pfedprompt} addresses personalization challenges in federated vision-language models. ``Personalized Prompt for Vision-Language Models in Federated Learning" investigates how to learn personalized prompts. These prompts are tailored to individual clients or datasets within the federated system. The aim is to unlock performance gains by having the model adapt its behavior for specific local data distributions.
    \item \textit{FedTPG:}  (Text-driven Prompt Generation for Vision-Language Models in Federated Learning) \cite{qiu2024federated} aims to enhance prompt generation techniques in a federated context. It introduces the idea of learning a prompt generator network which can produce context-aware prompts that guide vision-language models to tackle a variety of tasks. This has potential benefits for scenarios where a model must adapt to new classes or data it hasn't encountered previously, aligning well with the distributed nature of federated learning.
    \item \textit{FedMM:} In computational pathology, fusing information from multiple modalities can significantly improve diagnostic accuracy. However, centralized training approaches raise privacy concerns due to the sensitive nature of medical images. FedMM introduces a federated framework designed specifically to handle multi-modal data in this context.
    The key idea of \cite{peng2024fedmm} is to train individual feature extractors for each modality in a federated manner. Because only these learned feature extractors are shared, raw image data remains protected within each institution.  FedMM can accommodate the situation where different institutions or hospitals may have different sets of available modalities. It enables collaborative learning even with this data heterogeneity. Subsequent tasks like classification can be performed locally using the features extracted by the federated models.
    \item \textit{FedDAT:} Foundation models offer impressive performance across many tasks but often require substantial amounts of data for finetuning.  FedDAT \cite{chen2024feddat} addresses the challenge of finetuning these models in a federated context where the goal is to protect data privacy. To handle heterogeneity, FedDAT leverages a Dual-Adapter Teacher technique to regularize how model updates are made on each client.  Furthermore, it employs Mutual Knowledge Distillation to facilitate efficient knowledge transfer across clients in the federated system. 
    \item \textit{CLIP2FL:}  Real-world data is often messy, and client devices in a federated system  might have data with different characteristics or class imbalances. CLIP2FL \cite{shi2024clip} tackles this by using a pre-trained CLIP model as guidance. On the client-side, CLIP is used for knowledge distillation to improve the local feature representations. On the server-side, CLIP is employed to generate features which help retrain the server's classifier, mitigating the negative impact of the long-tailed data problem. 
    \item \textit{FedAPT:} \cite{su2024federated} introduces FedAPT, a novel method for collaborative learning in federated settings where data resides on multiple clients with varying domains (e.g., different image styles or categories). FedAPT aims to improve model generalization across these domains while maintaining data privacy. The key innovation lies in adaptive prompt tuning within the federated learning framework. Instead of directly sharing raw data, FedAPT trains a meta-prompt and adaptive network to personalize text prompts for each specific test sample. This allows the model to better adjust to domain-specific characteristics.
    \item \textit{General Commerce Intelligence:} \cite{lee2024general}  discusses the development of a novel NLP-based engine designed for commerce applications. This engine leverages federated learning to provide personalized services while ensuring privacy preservation across multiple merchants. The authors focus on creating a ``glocally" (globally and locally) optimized system that balances global optimization needs with local data privacy requirements.
\end{itemize}

\subsection{Practical applications on Federated Foundation Models}

Training foundation models within a federated learning framework presents distinct challenges, particularly due to the disparate nature of data sources and the varied computational resources across participating devices. The overarching goal is to cultivate effective and inclusive training strategies that can efficiently manage device heterogeneity and ensure data privacy, all while maintaining high model performance.

Training foundation models from scratch within a federated learning context is an ambitious endeavor that involves complex coordination and robust algorithmic strategies. Unlike traditional centralized training environments, federated learning necessitates handling data that remains on local devices, preventing the direct sharing of raw data. This scenario demands sophisticated techniques to efficiently aggregate learnings from disparate data sources, which are often uneven in size and diversity. The primary challenge lies in ensuring that the model learns effectively from each node without requiring extensive computational resources or compromising the integrity and privacy of the data. To overcome these hurdles, training strategies must be carefully designed to optimize the learning process across the network, allowing for both model convergence and performance retention. Such strategies often involve advanced algorithms for secure multi-party computation, differential privacy, or decentralized optimization methods. By training foundation models from scratch in this way, the federated approach not only safeguards data privacy but also harnesses the unique insights embedded in local data distributions, leading to more robust and generalizable models.

Prompt learning is emerging as a pivotal approach in both natural language processing and computer vision fields, enabling models to adapt to new tasks with minimal changes to their architecture or weights. This section explores the integration of prompt learning with federated learning (FL) across different domains, highlighting recent advancements and unique applications.

Furthermore, beyond merely fostering participation, it is crucial to consider how profits and costs associated with deploying FMs via APIs are distributed. Ensuring a fair allocation of rewards and benefits is imperative to maintain trust and promote sustained cooperation among stakeholders. Mechanisms need to be established to define the distribution of profits derived from the use of FMs, guaranteeing a fair share of economic benefits. This equitable distribution is essential not only for fostering a sense of fairness but also for encouraging continued participation and investment in the FL ecosystem for FMs.

The concept of Federated Foundation Models is at the forefront of federated learning, enabling the training of large-scale models across distributed networks. This methodology is particularly effective in dealing with the challenges related to synchronizing and updating model parameters in environments where data quality and quantity are inconsistent across nodes. It ensures that learning is continuous and effective, even when network conditions and data availability vary significantly \cite{yu2023federated}.

Additionally, the work titled ``Heterogeneous Ensemble Knowledge Transfer for Training Large Models in Federated Learning" by Cho et al. \cite{cho2022heterogeneous} explores innovative techniques for transferring knowledge in federated settings. This study is crucial for the development of robust models capable of performing well across diverse network conditions. By facilitating knowledge transfer, this approach allows for the aggregation of insights from different data distributions and device capabilities, which is essential for building comprehensive and resilient models.

Furthermore, ``No One Left Behind: Inclusive Federated Learning over Heterogeneous Devices" by Liu et al. \cite{liu2022no} focuses on creating federated learning algorithms that integrate every participating device, regardless of its computational capabilities or the quality of the data it holds. This inclusivity ensures that every device contributes to and benefits from the collaborative learning process, thus maximizing the utilization of available data and enhancing the overall performance of the model. This approach is fundamental to achieving equity in model training and ensuring that the advantages of sophisticated model learning are universally accessible.

These studies provide a foundation for further research into strategies that enhance the combination of foundation models in federated learning frameworks, prioritizing inclusivity and efficiency.

\section{Foundation Models in Biomedical Healthcare}\label{sec:fm-medical}
This section is dedicated to a comprehensive exploration of the application of foundation models within the biomedical healthcare domain, focusing on both language and vision-language models. It will delve into the benchmarks and setups employed to evaluate these models, highlighting the specialized frameworks and metrics used to assess their performance in healthcare-specific downstream tasks in foundation models. 

\subsection{Biomedical Foundation Models}
The application of foundation models has revolutionized numerous fields, particularly in natural language processing (NLP) and vision-language multimodal tasks. This transformation is largely attributed to several pivotal factors. Firstly, extensive pre-training on large text corpora allows these models to develop comprehensive universal language representations, which significantly enhance performance on various downstream tasks \cite{qiu2020pre}. Secondly, such pre-training provides an improved initialization for models, which not only boosts generalization capabilities but also speeds up convergence on specific target tasks. Thirdly, this method acts as a powerful form of regularization, crucial for preventing overfitting, particularly when training data is scarce. 

Meanwhile, Vision-language multimodal models \cite{yuan2021florence, radford2021learning, liu2023medical} are emerging as a powerful subset of foundation models, particularly in the field of image classification tasks in biomedical healthcare. These models synergistically combine the capabilities of image processing and language understanding to tackle complex tasks that require the integration of visual and textual data. In the healthcare sector, this ability is invaluable, as it enables the models to interpret medical imagery, such as scans and X-rays, alongside associated clinical notes or diagnostic information \cite{litjens2017survey}. For example, a vision-language model might analyze an MRI scan while simultaneously considering a patient’s written medical history to provide a more accurate diagnosis. This dual capability enhances the model's precision in identifying disease markers, understanding patient symptoms, and suggesting appropriate medical interventions. The integration of these two modalities in a single model not only streamlines the diagnostic process but also improves the accuracy of treatment recommendations, paving the way for more personalized and effective healthcare solutions. By leveraging these advanced models, medical professionals can gain deeper insights into patient conditions, leading to better outcomes and more efficient management of healthcare resources.

The training of foundation models (FMs) in the biomedical domain involves several crucial phases that enhance their applicability and effectiveness. Initially, unsupervised pretraining \cite{caron2019unsupervised} plays a pivotal role, where models learn from large corpora without labeled data. This phase emphasizes the discovery of inherent structures and abstract relationships within the data, without the need for specific predictive tasks, making it invaluable for identifying complex patterns. Subsequently, self-supervised learning forms the backbone of foundation models, traditionally utilizing unstructured text from general-domain sources such as Wikipedia or web-crawled pages. Recent advancements, however, have steered the customization of pre-trained FMs towards specific fields to better meet domain-specific requirements. For instance, CodeBERT \cite{feng2020codebert} is meticulously trained on programming languages to proficiently comprehend and generate code, whereas SciBERT \cite{zajkac2023clinician} is tailored for parsing scientific publications and biological sequences, addressing the unique challenges of academic and medical research. Following this, reinforcement learning from human feedback (RLHF) \cite{christiano2017deep} introduces a novel fine-tuning approach where models are adjusted based on rewards derived from human feedback rather than traditional labels. This method significantly aligns model outputs with human values and preferences, essential for applications demanding high engagement and accuracy in user interactions. Lastly, in-context learning \cite{min2022rethinking}, especially effective in models like GPT, leverages the model's capacity to generalize from a few examples. By presenting models with specific examples of the desired task at inference time, they dynamically adapt their responses to the context, enhancing their flexibility and utility without the need for additional training. This sequence of training methods collectively enhances the adaptability and performance of FMs, making them highly suitable for sophisticated tasks in the biomedical domain.

Foundation models can greatly benefit from training on expanded, domain-specific corpora \cite{gu2021domain}. For achieving peak performance in specialized downstream tasks, it is increasingly recognized that integrating in-domain data during the training phase is imperative. This targeted approach not only refines the model’s understanding of complex biomedical terminologies but also significantly enhances its practical applications in healthcare. By tailoring the training process to incorporate specific biomedical vocabulary and contextual nuances, FMs can be transformed into more effective tools, offering substantial improvements in processing and understanding medical texts, which is vital for advancing innovations and solutions within the healthcare industry.

\subsection{Biomedical Healthcare ML Applications and Benchmarks}

\paragraph{Applications}
The application of FMs in the biomedical domain is propelled by a range of compelling reasons, each underscoring the unique challenges and opportunities this field presents.

\begin{itemize}
    \item \textit{Complexity of Sequential Biomedical Data:} Biomedical information, including electronic health records and biomedical texts, often comes in the form of sequential tokens lacking annotations. Historically, this complexity posed significant hurdles for effective data modeling. However, advancements in FMs have enabled effective training on such data in a self-supervised manner, significantly expanding the possibilities for processing and understanding biomedical information using these sophisticated models.
    \item \textit{Scarcity of Annotated Data:}  In the biomedical field, annotated data is typically scarce and expensive to produce, often leading to ``zero-shot" or ``few-shot" learning scenarios. Recent developments in language models, notably GPT-3 \cite{brown2020language}, have showcased remarkable capabilities in few-shot and even zero-shot learning. This evolution means that a well-trained FM can act as a powerful feature extractor in the biomedical domain, reducing the dependency on large volumes of annotated data and easing the barriers to entry for complex biomedical analysis.
    \item \textit{Knowledge Intensity:} The biomedical sector is densely packed with specialized knowledge, much more so than general domains, often necessitating expert-level understanding. FMs serve as an accessible, soft knowledge base \cite{petroni2019language}, which can assimilate and replicate expert knowledge from vast biomedical texts without direct human annotation. For example, GPT-3 has shown an impressive ability to recall and apply extensive, intricate common knowledge in practical applications \cite{brown2020language}, demonstrating its utility as a tool for knowledge dissemination and decision support in healthcare.
    \item \textit{Diversity of Biological Data:} The scope of biomedical data extends beyond textual information to include diverse biological sequences, such as proteins and DNA. The application of FMs to these types of data has been notably successful, particularly in tasks like protein structure prediction. This success underlines the potential of FM to tackle a broader array of biological challenges, suggesting a promising future where FMs contribute substantially to critical tasks in genomics, proteomics, and other areas of biological research.
    \item \textit{Speed of Knowledge Synthesis \cite{saxena2022large}:} The rapid pace at which biomedical knowledge evolves makes it challenging to keep up with the latest research and clinical practices manually. FMs, trained on the latest corpus of literature and clinical guidelines, can quickly synthesize new information, making them invaluable tools for healthcare professionals who need to stay informed about the latest developments in real-time.
    \item \textit{Enhanced Predictive Analytics \cite{gupta2021cross}:} FMs have the potential to revolutionize predictive analytics in healthcare by integrating diverse data types—from patient records to research articles—to predict disease outbreaks, patient outcomes, and treatment efficacy. This capability can lead to more personalized medicine, where treatments are tailored to individual patients based on predictions made by these models.
    \item \textit{Automated Reasoning and Decision Support:} FMs can be employed to automate reasoning processes and support decision-making in clinical environments. By processing and analyzing large volumes of medical data, these models can suggest diagnostic options, propose treatment plans, and even predict possible complications, thereby assisting medical professionals in making better-informed decisions.
    \item \textit{Reduction in Diagnostic Errors \cite{horiuchi2024accuracy}:} By providing comprehensive, data-driven insights, FMs can help reduce diagnostic errors, one of the significant challenges in healthcare. Their ability to learn from vast datasets and identify patterns that may be overlooked by human experts can contribute to more accurate diagnoses and, consequently, more effective treatments.
\end{itemize}

These factors collectively motivate the integration of foundation model into biomedical research and healthcare operations, indicating a robust pathway for leveraging AI to manage and utilize complex biomedical data more effectively.

\paragraph{Benchmarks}
The application of pre-trained FMs in the biomedical field exploits a diverse array of unstructured data sources, including electronic health records, scientific publications, social media texts, biomedical image-text pairs, and various biological sequences such as proteins. For a comprehensive review of mining electronic health records (EHR), please refer to the previous survey \cite{eichelberg2005survey}. Discussions on the integration of health records and social media texts are explored in \cite{gonzalez2017capturing}, while a systematic overview of biomedical textual corpora is presented in \cite{kalyan2020secnlp}.

\textbf{Key Benchmarks in Biomedical Research:}
\begin{itemize}
    \item \textit{Electronic Health Records (EHR):} EHRs encapsulate a comprehensive digital record of patient information, including demographics, medical history, medications, laboratory test results, and billing details. They are pivotal for longitudinal studies, allowing researchers to track patient outcomes over time and identify patterns and predictors of diseases. The vast amount of data contained within EHRs makes them invaluable for training FMs to recognize and predict medical conditions accurately, although access is tightly regulated to protect patient privacy \cite{solares2020deep,weng2019representation}.    
    \item \textit{MIMIC-III (Medical Information Mart for Intensive Care III):} This critical care database contains detailed information from over 58,976 ICU admissions, including  2,083,180  vital signs, medications, laboratory measurements, observations, and notes. This richness makes MIMIC-III ideal for developing models that predict patient outcomes, tailor treatments, and conduct epidemiological studies in critical care settings \cite{johnson2016mimic}.
    \item \textit{CPRD (Clinical Practice Research Datalink):} A comprehensive dataset that provides a complete medical record from GP practices in the UK. It includes diagnoses, prescriptions, and clinical events, making it highly suitable for observational studies and clinical trials. The linkage to secondary care data enhances its utility in comprehensive healthcare research \cite{herrett2015data}.    
    \item \textit{Reddit and Tweets:} These datasets are increasingly used for public health monitoring and sentiment analysis. Reddit’s COMETA corpus and Twitter’s COVID-twitter-BERT provide real-time data on public health trends, misinformation patterns, and community response to health crises, which are crucial for understanding public health behavior and improving communication strategies \cite{basaldella2020cometa, muller2023covid}.    
    \item \textit{MIMIC-CXR:} This dataset of chest x-rays and accompanying radiological reports is crucial for developing automated diagnostic tools that assist radiologists in detecting and diagnosing pathologies from imaging studies. The textual descriptions help train models to correlate visual signs with diagnostic language \cite{johnson2019mimic}.
    \item \textit{DNA Dataset:} This genomic dataset facilitates the training of models on genetic sequences to predict gene functions, understand genetic variations, and assist in personalized medicine strategies. It is essential for advancing genomics research and integrating genetic information with clinical data \cite{ji2021dnabert}.
    \item \textit{FMRI datasets:} These datasets comprise data from functional magnetic resonance imaging (fMRI) studies, which are invaluable in providing detailed insights into brain activity. Utilized extensively in neuroscience, fMRI data helps researchers understand brain functions, diagnose neurological disorders, and predict outcomes of therapeutic interventions. Notable datasets like the Philadelphia Neurodevelopmental Cohort (PNC) \cite{satterthwaite2016philadelphia}, Autism Brain Imaging Data Exchange (ABIDE) \cite{heinsfeld2018identification}, and UK Biobank \cite{alfaro2018image} include both functional and structural brain imaging data. These resources are critical for advancing our understanding of the brain, enhancing the accuracy of neurological diagnoses, and improving the efficacy of treatments by enabling a deeper analysis of the brain's response to various stimuli and conditions.
    \item \textit{The Human Protein Atlas:} Contains high-resolution images detailing the spatial distribution of proteins in human tissues and cells \cite{digre2023human}. This atlas is used for bioinformatics studies that integrate protein expression with gene expression data to elucidate cellular functions and disease mechanisms.
    \item \textit{GEUVADIS RNA sequencing dataset \cite{frazee2015polyester}:} Provides RNA sequencing data from multiple populations, which is crucial for understanding how genetic variation affects gene expression across different human populations. This dataset is instrumental in studying population genetics, evolutionary biology, and disease susceptibility.
    \item \textit{ImageCLEFmed\cite{pelka2020overview}:} A benchmark dataset for multimodal biomedical information retrieval that includes medical images, captions, and text descriptions. It supports tasks such as medical image classification, annotation, and retrieval, which are crucial for medical informatics applications.
\end{itemize}

\begin{table}[h!]
    \centering
    \begin{adjustbox}{width=0.9\columnwidth,center}
        \begin{tabular}{l|l|l}
        \toprule
        \textbf{Dataset} & \textbf{Size} & \textbf{Types} \\ \midrule
        MIMIC-III & 58,976 admissions & Text, Numeric, Categorical \\ 
        CPRD & 11.3 million patients & Text, Numeric, Categorical \\ 
        Reddit and Tweets & 800K reddit posts and up-to-date tweets & Text \\ 
        MIMIC-CXR & 77,110 images & Images, Text \\ 
        DNA Dataset & 106 DNA sequences & Genetic Sequences \\ 
        PNC & 9,500 participants & Imaging (Functional, Structural) \\
        ABIDE & Over 1,100 individuals & Imaging (Functional, Structural) \\ 
        UK Biobank & Over 500,000 participants & Imaging (Functional, Structural), Genetic, Text \\ 
        Human Protein Atlas & 12,003 proteins & Images, Text \\
        GEUVADIS RNA sequencing & 462 individuals & Genetic Sequences \\ 
        \bottomrule
        \end{tabular}
    \end{adjustbox}
    \caption{Overview of Key Biomedical Healthcare Benchmarks.}
    \label{tab:biomedical_datasets}
\end{table}

These datasets exemplify the diverse types of biomedical data available for research, each offering unique insights and challenges that can be leveraged to train more effective and nuanced FMs for varied applications in healthcare and medical research. Note that the numerical details of the datasets are demonstrated in Table.~\ref{tab:biomedical_datasets}.

\subsection{Biomedical Healthcare on Large Language Models}
As introduced in Section.~\ref{subsec:backbone}, the backbone of most pre-trained foundation models, including prominent ones like BERT, GPT, T5, and their variants, is founded on the Transformer architecture, which framework is characterized by its reliance on self-attention networks and feed-forward networks (FFNs). The benign enables dynamic interactions between tokens, enhancing the model’s ability to handle complex input relationships, while FFNs perform non-linear transformations to deepen token representations, bolstering feature extraction capabilities.

In parallel, the evolution of text representation through FMs has significantly advanced from initial static word embedding methods to sophisticated models capable of understanding contextual nuances \cite{wang2020positiona,wang2019encoding}. Historical neural language models laid the groundwork by predicting word contexts in a unidirectional manner, but modern approaches like ELMO \cite{peters2018deep}, GPT \cite{radford2018improving}, and BERT \cite{devlin2018bert} have transformed the landscape with bi-directional and context-aware strategies. These models, through methodologies such as bidirectional language modeling and masked language model tasks, offer dynamic, context-sensitive word representations that vastly enhance performance across diverse NLP applications, making them fundamental to contemporary language processing tasks.

\subsubsection{How to Tailor LLMs to the Biomedical Domain}
The adaptation of large language models to the biomedical domain involves specialized methodologies tailored to enhance their functionality for this sector's unique tasks. Initially crafted for general natural language processing (NLP) tasks, models like BERT \cite{devlin2018bert} typically undergo a two-stage training process: initial training through a self-supervised meta-task (such as a masked language model or causal language model) on a broad, task-agnostic corpus, followed by fine-tuning on more specialized, often smaller-scale, downstream tasks relevant to specific fields. The two strategies have been developed to better integrate LLMs into the biomedical field are as follows:

\begin{itemize}
    \item \textit{Continual Pre-training:} This method involves taking general LLMs such as BERT, initially pre-trained on extensive general corpora like Wikipedia or BookCorpus, and continuing their training on domain-specific corpora, such as PubMed texts and MIMIC-III data. For instance, BioBERT \cite{lee2020biobert} extends BERT's training to include PubMed abstracts and articles, while BlueBERT \cite{peng2019transfer} is further trained on both PubMed and MIMIC-III texts. These adaptations often retain the original model’s vocabulary, which may not fully capture the specialized terminology of biomedical texts \cite{gu2021domain}.
    \item \textit{Pre-training from Scratch:} Some research advocates starting anew with domain-specific corpora to tailor PLMs more closely to biomedical needs \cite{beltagy2019scibert, gu2021domain}. SciBERT \cite{beltagy2019scibert} is an example of this approach, where a novel vocabulary of 30,000 terms specific to the domain was developed, and the model was trained on a corpus comprising both computer science (18\%) and biomedical (82\%) texts. However, recent findings suggest that mixed-domain pre-training might not be optimal for applications requiring high domain specificity. Instead, exclusive pre-training on biomedical corpora is recommended to ensure maximum relevance and efficacy.
\end{itemize}

\paragraph{Related Literatures} 
Before exploring the applications of Large Language Models in the biomedical healthcare domain, it is essential to recognize several representative surveys and peer-reviewed publications that have thoroughly reviewed the landscape of biomedical language models. These resources provide invaluable insights into the development, applications, and future prospects of LLMs within this specialized field, laying a foundational understanding for ongoing and future research. Several key surveys and publications have extensively discussed the current state and potential advancements of transformer-based biomedical models and general prompting methods in natural language processing:

\begin{itemize}
    \item \textit{AMMU: A Survey of Transformer-Based Biomedical Pretrained Language Models \cite{kalyan2022ammu}:} This comprehensive survey examines the evolution and impact of transformer-based models that have been specifically developed for the biomedical field. The survey details various approaches to adapting general language models to address the unique challenges posed by biomedical texts, such as the high specificity of vocabulary and the critical nature of the accuracy needed in medical contexts. The authors discuss multiple models that have been successfully implemented, highlighting their methodologies, the datasets they were trained on, and their performance on different biomedical NLP tasks. It provides a critical analysis of the strengths and limitations of these models, offering insights into how the field might evolve and suggesting directions for future research to enhance model accuracy and applicability.    
    \item \textit{Pre-train, Prompt, and Predict: A Systematic Survey of Prompting Methods in Natural Language Processing \cite{liu2023pre}: } by Liu et al. \cite{liu2023pre}: This survey explores the relatively new technique of prompting, which adapts pre-trained models to specific tasks using minimal task-specific data. Prompting involves modifying the input to pre-trained models in such a way that the task is reformulated to leverage the model's existing knowledge. The survey systematically categorizes different types of prompts, discusses their applications in various NLP tasks, and evaluates their effectiveness across several benchmarks. It provides a detailed look at how prompting can reduce the need for large annotated datasets, which is particularly beneficial in domains like biomedicine where acquiring such data can be costly and time-consuming. The paper also considers the future of prompting in NLP, suggesting that further refinement of prompting strategies could lead to more generalizable and efficient NLP systems.    
    \item \textit{Foundation Models in Healthcare: Opportunities, Risks \& Strategies Forward \cite{thieme2023foundation}:} This survey delves into the dual-edged nature of applying foundation models within the healthcare sector. It discusses the substantial opportunities these models present, such as enhancing diagnostic accuracy, predicting patient outcomes, and personalizing treatment plans. However, it also addresses the significant risks involved, particularly concerning data privacy, model bias, and the ethical implications of automated decision-making in healthcare. The authors propose a framework of strategies to mitigate these risks while capitalizing on the potential benefits. These strategies include developing robust governance frameworks, ensuring transparency in model workings, and engaging with a broad range of stakeholders to ensure that the deployment of these models in healthcare settings is both ethical and effective.    
    \item \textit{On the Opportunities and Risks of Foundation Models \cite{bommasani2021opportunities}:} This broad survey provides an extensive overview of the application of foundation models across various domains, with a particular focus on their transformative potential and the risks they pose. In the context of healthcare, the survey highlights how these models can revolutionize medical research and practice by providing new insights into disease patterns and patient care strategies. However, it also raises critical concerns about the reliability, fairness, and transparency of these models, especially given their potential to impact patient outcomes directly. The paper calls for a balanced approach to harnessing the power of foundation models, advocating for rigorous testing, ethical considerations, and regulatory oversight to ensure they benefit society as a whole.

\end{itemize}

\subsubsection{Practical LLMs in Biomedical Healthcare}
Since the introduction of BERT, a variety of biomedical pre-trained language models have been developed, enhancing the capabilities of NLP applications within the biomedical field. These models have been adapted either by further training on specialized in-domain corpora or by being built from scratch to cater specifically to the needs of medical and scientific communication. Below is a detailed summary of several existing pre-trained language models, where the specialized corpora, LLM backbone and released date are highlighted in Table.~\ref{tab:biomedical_llms}.

\begin{itemize}
    \item \textit{BioBert \cite{lee2020biobert}:} A pioneering work represents a significant advance in the application of language models to the biomedical domain. By adapting the BERT architecture, originally designed for general language understanding, BioBert is fine-tuned with biomedical texts sourced from extensive databases such as PubMed abstracts and PMC full-text articles. This adaptation is not merely a continuation of training but a targeted effort to align the model's learning with the intricacies and terminologies unique to biomedical literature. As a result, BioBert excels in several biomedical text mining tasks including named entity recognition, relation extraction, and question answering over biomedical knowledge bases. The strength of BioBert lies in its ability to capture deep semantic connections between biomedical concepts, significantly improving the model's utility for researchers and healthcare professionals who rely on swift and accurate interpretations of medical texts.
    \item \textit{MedBert \cite{rasmy2021med}:} MedBert is an innovative approach to creating a language model that is steeped from the outset in the medical context. Unlike models that are adapted from general-purpose architectures, MedBert is pre-trained from scratch on a large and diverse corpus of medical texts, including electronic health records and other clinical documents. This ground-up approach allows MedBert to develop a nuanced understanding of medical language, including jargon, abbreviations, and the complex relationships between medical concepts. The model has shown significant improvements in tasks such as patient phenotyping and diagnostic prediction, making it a vital tool for healthcare analytics. MedBert's design addresses the challenges of applying general language models to medical data, ensuring that the nuances and critical details of medical communication are not lost in translation.
    \item \textit{ClinicalBERT \cite{huang2019clinicalbert}:} Tailored for understanding and processing clinical notes, ClinicalBERT was trained exclusively on data from the MIMIC-III database \cite{johnson2016mimic},, which includes around 2 million clinical notes. This specialized training prepares ClinicalBERT to handle a variety of clinical documentation styles and medical shorthand, making it an invaluable tool for applications like patient outcome prediction and automated documentation review, which require a deep understanding of clinical narratives.
    \item \textit{SciBERT \cite{beltagy2019scibert}:} Developed from scratch, SciBERT focuses on scientific text, primarily from the biomedical field, leveraging a corpus of papers available through the Semantic Scholar database. With 82\% of its training corpus composed of biomedical research articles, SciBERT is adept at deciphering complex scientific terminology and extracting relevant information from scholarly articles, thereby facilitating advanced text mining and information retrieval tasks in scientific research.
    \item \textit{COVID-twitter-BERT \cite{muller2023covid}:} This model was specifically developed to analyze and understand discourse about COVID-19 on Twitter. It was trained during the initial stages of the pandemic on a dataset comprising approximately 160 million tweets related to the virus. The model is designed to capture the nuances of public sentiment, misinformation, and evolving topics related to COVID-19, providing valuable insights for public health officials and researchers studying communication patterns during health crises.
    \item \textit{MedGPT \cite{kraljevic2021medgpt}:} Inspired by the GPT architecture, MedGPT was trained on electronic health records (EHRs) and is designed to predict future medical events based on patients’ medical histories. Its training allows it to model and predict various outcomes, such as diagnoses and complications, making it a potential tool for prognostic assessments in clinical settings.
    \item \textit{SCIFIVE \cite{phan2021scifive}:} This model is a domain-specific adaptation of the T5 model, trained under the Seq2seq framework on extensive biomedical corpora. SCIFIVE is engineered to transform complex biomedical queries into concise answers, facilitating tasks such as summarizing scientific texts and generating explanatory notes from dense medical data.
    \item \textit{LLMBiomedicine \cite{monajatipoor2024llms}:} This research highlights the effectiveness of meticulously designed prompts and the strategic selection of in-context examples to enhance the performance of LLMs on biomedical NER tasks. By adjusting prompts and examples to better fit the context of biomedical data, the study demonstrates significant improvements in model performance, making LLMs more adept at identifying and classifying medical entities in text.
    \item \textit{ClinicalGPT \cite{wang2023clinicalgpt}:} ClinicalGPT, a model that has been fine-tuned with a diverse set of medical data to enhance its performance and reliability in clinical scenarios. The model undergoes rigorous evaluations to ensure it meets the high standards required for medical applications, focusing particularly on its ability to maintain factual accuracy and provide contextually appropriate responses in simulated clinical interactions. ClinicalGPT represents a significant advancement in the use of LLMs in medicine, offering potential improvements in automated patient interaction, diagnostic support, and personalized treatment planning. By leveraging a vast corpus of medical texts for fine-tuning, the model is better equipped to handle the nuanced and highly specialized language found in clinical notes and patient interactions.
    \item \textit{MultiMedQA \cite{singhal2023large}:} MultiMedQA is a comprehensive benchmark combining six existing medical question answering datasets, which span a variety of contexts from professional medicine to consumer health inquiries. The benchmark is enhanced by a newly developed dataset, HealthSearchQA, which consists of medical questions frequently searched online. This diverse collection of datasets is utilized to test the LLMs' ability to understand and process complex medical information across different facets of healthcare and patient inquiries. The authors discuss the significant challenge of assessing LLMs in clinical settings, where the accuracy of information and the models' understanding of nuanced medical language are crucial. By employing MultiMedQA, the authors aim to provide a more nuanced and thorough evaluation of LLMs than previous benchmarks allowed. 
    \item \textit{Chatdoctor \cite{yunxiang2023chatdoctor}:} Chatdoctor improves the performance and relevance of responses in medical conversational systems. To achieve this, the model was fine-tuned using a substantial dataset of 100,000 real-world patient-physician conversations sourced from online medical consultations. This approach ensures that the model not only understands medical terminology and procedures but also grasps the nuances of patient interactions and inquiries.
    \item \textit{Taiyi \cite{luo2024taiyi}:} Taiyi highlights the limitations of existing fine-tuned biomedical LLMs, which are predominantly monolingual and focused on question answering and conversation tasks within the biomedical field. Taiyi, by contrast, is designed to enhance performance across a broader spectrum of NLP applications, including entity extraction, relation extraction, and information retrieval, catering to both English and non-English texts. The development and evaluation of Taiyi involve rigorous fine-tuning processes that adjust the model to grasp the nuances and specific terminology used in various biomedical contexts, significantly improving its utility and applicability in a global healthcare context. This model represents a substantial advancement in the field of biomedical NLP by supporting multilingual capabilities and addressing the critical need for diverse language processing in medical research and healthcare delivery.
\end{itemize}

\begin{table}[t!]
    \centering
    \begin{tabular}{l|l|l|l}
    \toprule
    \textbf{Model Name} & \textbf{Corpora} & \textbf{LLM Backbone} & \textbf{Release Date} \\ \midrule
    BioBert & PubMed abstracts, PMC articles & BERT & 2020 \\ 
    MedBert & Medical texts, EHRs & BERT & 2021 \\ 
    ClinicalBERT & MIMIC-III clinical notes & BERT & 2019 \\ 
    SciBERT & Scientific papers (82\% biomedical) & BERT & 2019 \\ 
    COVID-twitter-BERT & Tweets about COVID-19 & BERT & 2023 \\ 
    MedGPT & Electronic health records (EHRs) & GPT & 2021 \\ 
    SCIFIVE & Biomedical corpora & T5 & 2021 \\ 
    LLMBiomedicine & Biomedical texts (NER  \cite{sang2003introduction} tasks) & GPT-4 & 2024 \\ 
    ClinicalGPT & Diverse medical data & GPT & 2023 \\ 
    MultiMedQA & Medical QA datasets & PaLM \cite{chowdhery2023palm} & 2023 \\ 
    Chatdoctor & Patient-physician conversations & LLaMa & 2023 \\ 
    Taiyi & Biomedical texts, multilingual & Qwen \cite{bai2023qwen} & 2024 \\ 
    \bottomrule
    \end{tabular}
    \caption{Overview of Pre-trained Language Models in Biomedicine with Release Dates}
    \label{tab:biomedical_llms}
\end{table}

\subsection{Biomedical Healthcare on Vision Language Models}
This section elaborates on the training of vision language models for biomedical imaging, and their practical applications in the biomedical healthcare sector.

\subsubsection{How to Train Vision Language Models for Biomedical Imaging}
Deep neural networks demonstrate outstanding performance in various vision tasks, including image classification, object detection, and instance segmentation. A key to this success in the foundation model era is the concept of pre-training, which, unlike in NLP where it usually involves language models, traditionally meant training on extensive labeled image datasets like ImageNet \cite{deng2009imagenet}. More recently, diverse learning methods have been introduced to overcome limitations of conventional supervised learning, such as generalization errors and spurious correlations. We examine several methodologies suitable for imaging applications as follows:

\begin{itemize}
    \item \textit{Unsupervised Pre-training:} Unsupervised pre-training leverages large volumes of unlabeled image data to learn rich feature representations without the guidance of explicit annotations. Techniques such as autoencoders \cite{vincent2008extracting} and generative adversarial networks (GANs) \cite{goodfellow2014generative} train models to generate or reconstruct images, enabling them to capture the underlying data distributions and learn complex patterns within the visual inputs. This approach is particularly useful in domains where labeled data is scarce or expensive to obtain.
    \item \textit{Contrastive Self-supervised Learning:}  Contrastive self-supervised learning techniques \cite{chen2020simple,grill2020bootstrap,he2020momentum} train models to differentiate between various modifications of a given input image, such as determining whether two images are rotated versions of each other or entirely distinct. This method enables the model to develop features applicable to diverse vision tasks, including object detection and semantic segmentation.
    \item \textit{Masked Self-supervised Learning:} Drawing inspiration from BERT’s approach in NLP, masked self-supervised learning \cite{bao2021beit,he2022masked,xie2022simmim} is gaining popularity in computer vision. This generative pre-training method trains models to reconstruct images from partially obscured inputs, aiding in understanding the underlying structure of visual data.
    \item \textit{Contrastive Language-image Pre-training:} An innovative method, contrastive language-image pre-training \cite{radford2021learning} (CLIP), involves training a vision model using diverse image-text datasets. The model learns to match images with corresponding texts within a mini-batch through contrastive learning. CLIP shows impressive zero-shot capabilities, performing on par with traditional models like ResNet \cite{he2016deep} on ImageNet without task-specific training. Text descriptions enhance understanding of the visual content, facilitating the model’s comprehension of visual elements and their interrelations, which is essential for effective learning.
    \item \textit{Instructed fine-tuning:} Instructed fine-tuning involves explicitly guiding the model during the fine-tuning process with task-specific instructions. This method builds upon the foundation established during pre-training by aligning the model's learning objectives closely with the nuances of the target task \cite{radford2021learning}. For example, in biomedical imaging, models can be instructed to identify specific medical conditions from images using detailed descriptions of symptoms or expected imaging features. This approach helps the model to focus on relevant aspects of the data, enhancing its performance on specialized tasks such as diagnosing diseases from medical scans.
\end{itemize}

Note that a significant challenge in harnessing FMs for vision-language tasks lies in overcoming the ``task gap" and the ``domain gap." The task gap refers to the differences between the generic meta-tasks used in FMs, such as masked language modeling in BERT or causal language modeling in GPT, and the specialized requirements of downstream vision-language tasks, such as medical image annotation or diagnostic interpretation. The domain gap further highlights the disparity between the general training corpora used, and the highly specialized datasets needed for tasks in specific fields like biomedicine. To effectively deploy a pre-trained language model in vision-language applications within a specific domain, it is crucial to undertake both domain and task adaptations \cite{gu2021domain, gururangan2020don, rongali2020continual, zhang2020multi}. Domain adaptation involves additional training of a model—originally pre-trained on broad, general datasets within a targeted domain, such as biomedicine. This step ensures that the model becomes attuned to the specific terminologies and data types characteristic of the domain.

\subsubsection{Practical VLMs in Biomedical Healthcare}
Biomedical vision-and-language models have largely been shaped by influential self-supervised pre-training techniques, such as SimCLR \cite{chen2020simple} in computer vision and BERT \cite{devlin2018bert} in natural language processing. These foundational approaches have paved the way for the adoption of advanced text-to-image diffusion models \cite{ramesh2022hierarchical,rombach2022high,saharia2022photorealistic} in the medical field \cite{chambon2022roentgen,chambon2022adapting}, enhancing tasks ranging from diagnostic imaging to patient interaction. This subsection provides a detailed overview of the existing vision-and-language models (VLMs) within the biomedical sector and elucidates their functionalities. In this survey, VLMs in the biomedical healthcare sector are categorized into three primary types: dual-encoder, fusion encoder, and hierarchical structures. Each model type offers distinct advantages and limitations, tailored to specific application needs within the healthcare context.

Dual-Encoder Models process visual and textual inputs independently through separate encoders before merging the resulting vectors for final task execution. This architecture is particularly effective for tasks that require robust single-modal or crossmodal representation, such as image classification, image captioning, and cross-modal retrieval. However, the dual-encoder approach may fall short in fully capturing the intricate interplays between visual and linguistic elements, which can limit its effectiveness in more complex multimodal tasks.  Fusion-encoder models integrate visual and linguistic data early in the processing pipeline, utilizing a single encoder to manage both modalities. This method facilitates the capture of complex interactions between text and image, proving advantageous for tasks that demand a deep multimodal understanding, such as visual question answering and complex diagnostic reasoning. While fusion encoders excel in multimodal integration, they may encounter challenges in scenarios where a clear distinction between modalities is necessary. 

Besides the dual-encoder and fusion-encoder models, the field also explores innovative biomedical FMs that combine vision and language, such as hierarchical encoder alignment \cite{nguyen2019multi,gao2022pyramidclip} and medical text-to-image diffusion models \cite{saharia2022photorealistic,ruiz2023dreambooth}. Hierarchical alignment constructs input pyramids on both visual and linguistic sides, enhancing the model's ability to match features across modalities at multiple abstraction levels, which not only improves feature correspondence and model generalization but also optimizes the learning process, making it more efficient and adaptable to complex tasks like medical diagnosis from imaging and textual data. Such structured FMs offers significant advantages in terms of computational efficiency, robustness, and scalability, demonstrating potential for broad applications, especially in the biomedical field. Diffusion models are generative models inspired by non-equilibrium thermodynamics. They operate by defining a Markov chain of diffusion steps that gradually add random noise to data. The model then learns to reverse this diffusion process, reconstructing the desired data samples from the noise. This approach is particularly powerful in medical applications where generating high-fidelity images from textual descriptions can assist in diagnostic visualizations and treatment planning.

We summarize recent significant developments in adapting vision-language models for biomedical healthcare as follows, where the model type, encoder details, training corpora, and released dates are detailed at Table.~\ref{tab:VLM_biomed}.

\begin{table}[t!]
    \centering
    \begin{adjustbox}{width=0.95\columnwidth,center}
    \begin{tabular}{l|l|l|l|l|l}
    \toprule
    \textbf{Model Name} & \textbf{Type} & \textbf{Image Encoder} & \textbf{Text Encoder} & \textbf{Training Corpora} & \textbf{Release Date} \\ \midrule
    ConVIRT & Dual & ResNet & ClinicalBERT & MIMIC-CXR & 2022 \\ 
    GLoRIA & Dual & ResNet & BioClinicalBERT & Chexpert \cite{irvin2019chexpert} & 2021 \\ 
    MedCLIP & Dual & ResNet/ViT & BioClinicalBERT & Chexpert, MIMIC-CXR & 2022 \\ 
    CheXZero & Dual & CLIP-Image & CLIP-Text & Chest X-rays & 2022 \\ 
    LoVT & Dual & ResNet & ClinicalBERT & MIMIC-CXR & 2022 \\ 
    Adapted VLMs & Hierarchical & Diffusion, VAE \cite{doersch2016tutorial} & Bert, CLIP & Chexpert, MIMIC-CXR  & 2022 \\ 
    VisualBERT & Fusion & Varies & BERT & MIMIC-CXR & 2020 \\ 
    MedViLL & Fusion & ResNet & BERT & MIMIC-CXR & 2022 \\ 
    ARL & Fusion & CLIP-Image & RoBERTa \cite{liu2019roberta} & MedICaT \cite{subramanian2020medicat}, MIMIC-CXR, ROCO\cite{pelka2018radiology} & 2022 \\ 
    LViT & Fusion & ViT & BERT & QaTa-COV19 \cite{degerli2021reliable}, MoNuSeg \cite{kumar2017dataset} & 2023 \\ 
    RoentGen & Hierarchical & Diffusion & CLIP-Text & MIMIC-CXR & 2022 \\ 
    CLIPSyntel & Dual & CLIP & GPT-3.5 & MMQS \cite{ghosh2024clipsyntel} & 2024 \\ 
    Med-unic & Dual & ResNet/ViT & CXR-BERT \cite{boecking2022making} & MIMIC-CXR, PadChest \cite{bustos2020padchest} & 2024 \\ 
    EchoCLIP & Dual & ConvNeXt \cite{woo2023convnext} & CLIP-Text & Echocardiogram videos & 2024 \\ 
    Llava-med & Fusion & Llava \cite{liu2024visual} & Llava & PubMed \cite{zhang2023large}, PMC-15M \cite{zhang2023biomedclip} & 2024 \\ 
    \bottomrule
    \end{tabular}
\end{adjustbox}
    \caption{Overview of Vision-Language Models in Biomedical Healthcare.}
    \label{tab:VLM_biomed}
\end{table}

\begin{itemize}
    \item \textit{ConVIRT \cite{zhang2022contrastive}:} ConVIRT utilizes contrastive learning techniques to simultaneously train ResNet and BERT encoders on paired image and text data. This approach significantly improves the model's performance in image classification tasks by effectively reducing the dependency on large volumes of labeled data. By optimizing feature extraction and enhancing the semantic alignment between images and their textual descriptions, ConVIRT enables more accurate and efficient classification, making it particularly useful in scenarios where annotated datasets are limited.
    \item \textit{GLoRIA \cite{huang2021gloria}:} GLoRIA advances the field by employing both global and local contrastive learning strategies to finely align words in radiology reports with corresponding sub-regions within images. This method enhances local representation learning, allowing for more precise identification and classification of localized features in medical images. Such detailed alignment improves diagnostic accuracy and aids in the development of more sophisticated automated radiology analysis tools.
    \item \textit{MedCLIP \cite{medclip2022}:} MedCLIP leverages the innovative architecture of the CLIP model, specifically tailored for the medical domain. By utilizing pre-computed matching scores, MedCLIP enhances the alignment between medical images and their corresponding textual descriptions. This capability facilitates effective zero-shot learning, allowing MedCLIP to accurately classify medical conditions without the need for extensive fine-tuning on large annotated datasets. The model's ability to directly apply learned representations from diverse medical contexts makes it a valuable tool for rapid and efficient disease diagnosis, particularly in environments where labeled medical data is scarce.
    \item \textit{CheXZero \cite{tiu2022expert}:} CheXZero is another adaptation of the CLIP model, focused on zero-shot learning for medical imaging, specifically in chest radiography. Unlike traditional models that require detailed annotations for each new disease classification task, CheXZero applies the powerful zero-shot capabilities of the CLIP model to accurately identify pathologies in chest X-rays without additional model training. This approach is particularly beneficial for rapidly evolving medical scenarios, such as new disease outbreaks or rare conditions, where the availability of comprehensive labeled datasets might be limited. CheXZero's innovative use of CLIP for direct application in medical diagnosis demonstrates its potential to significantly streamline diagnostic processes in healthcare settings.
    \item \textit{LoVT \cite{muller2022joint}:} LoVT specifically targets localized medical imaging tasks by implementing a local contrastive loss that aligns representations of sentences or specific image regions. This alignment is crucial for tasks that require detailed understanding of small, localized anatomical structures or pathological features, enhancing the model’s accuracy in specialized medical imaging applications.
    \item \textit{Adapting Pretrained Vision-Language Foundational Models to Medical Imaging Domains \cite{chambon2022adapting}:} This study explores the effectiveness of adapting general vision-language models to the medical imaging domain. It demonstrates how foundational models, originally designed for broad applications, can be fine-tuned to meet the specific needs of medical diagnostics and research, thus broadening their applicability and improving performance in specialized tasks.
    \item \textit{VisualBERT \cite{li2020comparison}:} This study focuses on adapting general-domain vision-language models, such as LXMERT and VisualBERT, for the integration of medical images and texts. The effectiveness of these adapted models in disease classification showcases their potential in clinical settings, where they can support diagnostic processes and enhance the accuracy of medical assessments.
    \item \textit{MedViLL \cite{moon2022multi}:} MedViLL enhances the multimodal interaction between medical images and associated textual data through an innovative vision-language model framework by incorporating extensive medical knowledge and utilizing tailored masking schemes, MedViLL is specifically designed to improve understanding and generation tasks within the medical field, which excels in synthesizing comprehensive medical reports and generating detailed medical annotations, crucial for assisting healthcare professionals in making informed decisions. The approach to integrating complex medical datasets ensures a deeper contextual understanding and a more nuanced interpretation of both visual and textual medical data.
    \item \textit{ARL \cite{chen2022align}:} ARL (Align and Reasoning Language model) introduces a unique alignment strategy that specifically targets the challenges of medical imaging and text analysis. By aligning sentence or image region representations through a localized contrastive loss, ARL effectively bridges the gap between visual features and their corresponding textual annotations. This model is particularly adept at tasks that require precise localization of medical findings within images, supporting detailed diagnostic processes. ARL's focus on enhancing the correlation between detailed image regions and descriptive text makes it an invaluable tool for advanced medical imaging applications where accuracy and detail are paramount.  
    \item \textit{LViT \cite{li2023lvit}:} LViT leverages medical text annotations to significantly improve segmentation results, particularly in semi-supervised settings where labeled data may be scarce. By integrating rich textual information, LViT enhances its understanding of medical imagery, leading to more accurate segmentation and analysis of medical scans.
    \item \textit{RoentGen \cite{chambon2022roentgen}:} RoentGen introduces a pioneering approach by applying text-to-image diffusion models to medical imaging. This novel methodology holds promise for generating detailed and accurate medical images from textual descriptions, potentially revolutionizing the way medical imagery is produced and understood.
    \item \textit{CLIPSyntel \cite{ghosh2024clipsyntel}:} CLIPSyntel represents a synergistic application of CLIP and large language models to address the challenge of multimodal question summarization in healthcare. This model harnesses the strengths of both visual and textual data processing to provide concise and relevant summaries of complex medical inquiries, aiding healthcare professionals CLIP and LLM Synergy for Multimodal Question Summarization in Healthcare 
    \item \textit{Med-unic \cite{wan2024med}:} Med-unic presents an approach to enhance the performance of medical vision-language pre-training models across different languages. The authors focus on reducing bias in these models, which often perform better in languages with abundant training data (like English) compared to languages with less data. They introduce a unified framework that integrates multilingual textual features and visual content effectively. Their method involves using a debiasing technique that ensures more equitable learning from visual and textual data across various languages. This is achieved by carefully balancing the dataset and incorporating cross-lingual adaptation techniques to improve model performance uniformly across different linguistic contexts.
    \item \textit{EchoCLIP \cite{christensen2024vision}:} EchoCLIP is a specialized vision-language foundation model designed to improve echocardiography interpretation. EchoCLIP leverages the relationship between cardiac ultrasound images and expert cardiologist interpretations across diverse patient groups and diagnostic scenarios. The development of this model addresses the critical challenge of limited availability of annotated clinical data in cardiac imaging. By training on over one million cardiac ultrasound images, EchoCLIP aims to enhance the accuracy and efficiency of echocardiogram, offering a robust tool for cardiac diagnostics that can adapt to various clinical conditions and imaging indications.
    \item \textit{Llava-med \cite{li2024llava}:} Llava-med  presents a novel approach to training a vision-language conversational assistant tailored for the biomedical field. The study introduces a cost-efficient method for rapidly developing a multimodal conversational AI that can understand and discuss biomedical images alongside textual data. Unlike previous models that rely extensively on large-scale image-text pairs from general domains, Llava-med is trained specifically with biomedical data to better address the unique needs of the medical community.
\end{itemize}


\section{Open Challenges and Opportunities in Federated Foundation Biomedical Research}\label{sec:challenges}
The integration of AI technologies, particularly large pre-trained foundation models in the biomedical field, presents a range of future challenges and opportunities that must be tackled to unlock their full potential. This section delves into the key issues and potential avenues for progress concerning the application of federated foundation models in biomedical research. It underscores the need for robust solutions that ensure privacy, enhance model generalizability, improve computational efficiency, and address regulatory and ethical considerations. As we explore these challenges, we also highlight promising strategies that may pave the way for more effective and equitable AI-driven healthcare solutions.

\subsection{Challenges of Foundation Models in Biomedical Healthcare}
Research on foundation models in the biomedical healthcare domain presents several challenges and directions for future exploration:
\begin{itemize}
    \item \textit{Data Privacy and Security:} The primary challenge in foundation model, especially in healthcare, revolves around maintaining patient confidentiality and adhering to stringent data protection regulations like HIPAA \cite{act1996health} in the U.S and GDPR \cite{li2019impact} in Europe. Future research needs to focus on developing robust encryption methods and privacy-preserving algorithms that allow for the secure sharing of insights without exposing sensitive patient data \cite{challen2019artificial,chamikara2021privacy}.
    \item \textit{Model Generalization across Diverse Datasets:} FMs involve training models on highly heterogeneous data sources, often leading to challenges in model generalization. Research should explore techniques to enhance the generalizability of foundation models across diverse healthcare systems and varied patient demographics without compromising performance.
    \item \textit{Scalability and Computational Efficiency:} The computational demand for training large-scale LLMs and VLMs is significant. Optimizing resource allocation, reducing communication overhead, and efficient model updating mechanisms are crucial areas for future development to ensure scalability and practicality in real-world healthcare settings.
    \item \textit{Bias and Fairness:} Ensuring that foundation models do not perpetuate or amplify biases present in downstream tasks is critical \cite{wiens2019no,achiam2023gpt,chen2020treating}, especially under the biomedical domain, where the targeted problem for each patient can be narrow. Future research should include developing methodologies for bias detection and mitigation in model training and deployment phases. This also involves designing fair algorithms that provide equitable healthcare outcomes across different populations  \cite{kaushal2020geographic,zhao2020training}.
    \item \textit{Interoperability and Standardization:} There is a need for standardized protocols to ensure interoperability among different healthcare systems participating in foundation model learning. 
    \item \textit{Personalization:} Medical treatments and diagnostics often require high degrees of personalization. AI models must be capable of adapting to individual patient needs and conditions, which poses challenges in model design and data utilization without compromising generalizability.
    \item \textit{Scaling:} Deploying AI solutions on a large scale, particularly in diverse healthcare settings, presents logistical and computational challenges. Scalability involves not only the expansion of AI systems to handle larger datasets but also ensuring these systems are accessible across different regions and healthcare infrastructures.
    \item \textit{Biomedical Requirements of Accuracy:} Biomedical applications demand extremely high levels of accuracy and reliability. AI models used in diagnostics or treatment recommendation must meet rigorous standards to prevent errors that could adversely affect patient health.
    \item \textit{Robustness to Adversarial Attacks:} As the applications of a foundation model in biomedical scenarios can be distributed, they are susceptible to various types of adversarial attacks that can compromise model integrity. Enhancing the robustness of foundation models against such attacks, ensuring secure and reliable model performance, is a significant direction for ongoing research.
    \item \textit{Regulatory and Ethical Considerations:} As foundation models evolve, there will be increased scrutiny from regulatory bodies concerning their use in clinical settings. Research must address these regulatory challenges by developing models that are not only effective but also transparent and explainable to satisfy regulatory requirements and maintain public trust.
    \item \textit{Longitudinal Studies and Continuous Learning:} Implementing models that can adapt over time to new data and evolving biomedical conditions is crucial. Research into continuous learning mechanisms that allow FMs to update without forgetting \cite{li2017learning} previously learned knowledge while integrating new insights is essential for maintaining the relevance and accuracy of biomedical models.
\end{itemize}

\subsection{Opportunities in Federated Foundation Models}
Federated learning offers a unique framework for addressing several challenges associated with foundation models, particularly in the sensitive and data-intensive field of biomedical healthcare. In this survey, we would like to highlight how federated learning can help overcome the challenges of FMs and what opportunities it presents in both academic research and industrial applications:

\begin{itemize}
    \item \textit{Data Privacy and Security:} Federated learning enables the collaborative training of predictive models by sharing model updates rather than raw data. Each participating institution retains its data locally, significantly minimizing the risk of data breaches and unauthorized access. This method is especially beneficial in healthcare, where patient data is highly sensitive and subject to strict privacy regulations. Note that upholding data privacy is crucial not only for complying with laws like GDPR and HIPAA but also for maintaining patient trust. Federated learning's ability to train models without compromising data privacy helps healthcare organizations implement AI solutions without risking patient confidentiality or facing legal penalties. \textbf{Increased Model Robustness and Trustworthy:} Trust in medical AI systems is essential for their acceptance by both medical professionals and patients. Systems known for their reliability and backed by a transparent, accountable training process are more likely to be trusted and thus more widely adopted, which makes it crucial to further investigating the trustworthy federated foundation models. 
    \item \textit{Real-time Learning and Adaptation:} In federated learning frameworks, models can be updated continually as new data becomes available across the network. This dynamic learning process allows the models to adapt to emerging health trends or new strains of diseases. The ability to update and adapt foundation models in real-time is vital for keeping pace with the fast-evolving nature of diseases and treatments, ensuring that healthcare providers have the most current tools at their disposal.
    \item \textit{Collaborative Innovation:} By aggregating insights from diverse healthcare environments and patient demographics, federated learning facilitates the development of models that perform well across different settings. This heterogeneous data input helps the model learn more comprehensive patterns and reduces the risk of bias towards any particular group or condition. Specifically, how to use federated learning to efficiently establish a cooperative ecosystem where different healthcare entities can contribute to and benefit from shared AI advancements without compromising their data sovereignty is worth for real-world application, which can lead to more rapid development and refinement of AI technologies.
    \item \textit{Multimodality:} Medical data are inherently multimodal, encompassing an extensive array of data types—including text, images, videos, databases, and molecular structures—across various scales from molecules to populations \cite{kong2011integrative, ruiz2021identification}, and presented in both professional and lay language \cite{lavertu2019redmed,li2019neural}. While current self-supervised models excel within individual modalities, such as text \cite{lee2020biobert}, images \cite{chaitanya2020contrastive}, genes \cite{ji2021dnabert}, and proteins \cite{jumper2020high}, they typically lack the capability to integrate and learn from these diverse sources simultaneously. To truly leverage the vast and rich information available across different modalities, there is an urgent need to develop models that can perform both feature-level and semantic-level fusion. Successfully integrating these varied data types could revolutionize how biomedical knowledge is unified and significantly accelerate discovery processes in biomedicine. 
    Federated learning frameworks can capture a richer and more nuanced understanding of patient conditions, which is crucial for multimodal tasks like diagnosing complex diseases that may require correlating symptoms, radiology images, and genetic information.
    \item \textit{Synthetic Data Generation for Further Training:} Federated learning can facilitate the generation of synthetic training data, which helps address the scarcity of annotated datasets, particularly in specialized medical fields. By learning from diverse sources, federated models can generate new, synthetic examples that preserve the statistical properties of real data without revealing any individual patient's information. This synthetic data can then be used to further train and refine FMs across the network. The generation of synthetic data is a critical solution for overcoming data limitations in biomedical research, where privacy concerns and the rarity of certain conditions can significantly constrain the availability of training data. 
\end{itemize}

\section{Conclusions}\label{sec:conclusion}
This survey has delved into the transformative potential of foundation models and federated learning within the biomedical healthcare domains. Foundation models represent a significant advancement in artificial intelligence, offering robust, adaptable tools that can be fine-tuned for specific applications without constructing new models from scratch. These models, trained on expansive datasets, are capable of performing a wide array of tasks—from text generation to video analysis—that were previously beyond the reach of earlier AI systems. In the biomedical and healthcare sectors, where the efficacy of AI and the integrity of data privacy are crucial, foundation models play a pivotal role by enabling the extraction of valuable insights from constrained datasets. This survey has highlighted the current applications of FMs in these sectors, particularly focusing on large language models and vision-language models.

Meanwhile, Federated learning, characterized by its privacy-preserving and decentralized approach, complements the capabilities of foundation models perfectly. By combining the robust, generalizable nature of FMs with the privacy-centric, decentralized attributes of federated learning, researchers can perform deep analyses using globally pooled insights from locally held datasets. This synergy holds immense potential to meet the specific needs of biomedical AI applications, offering scalable solutions that accommodate the continuous updating of foundation models with new, relevant data.

Additionally, this survey has outlined various challenges and opportunities that arise with the adoption of federated foundation models in healthcare. Federated learning addresses critical issues such as data privacy, model generalization, scalability, and inherent biases within AI models. By allowing multiple institutions to collaboratively train models while keeping their data localized, federated learning not only complies with strict data privacy laws but also enhances the diversity and efficacy of medical AI applications. Key areas where federated foundation models could notably impact biomedical research and practice include enhancing model robustness and fairness, enabling real-time model updates and adaptations, and facilitating cross-institutional and international collaborations without compromising data security.

\bibliographystyle{unsrt}  
\bibliography{references}

\end{document}